\documentclass[electronics,article,accept,moreauthors,pdftex,10pt,a4paper]{mdpi} 

\usepackage[ruled,vlined]{algorithm2e}
\usepackage{cite}
\usepackage{amsmath}
\usepackage{amssymb}
\usepackage{amsfonts}
\usepackage{array}
\usepackage{color}
\usepackage{soul}
\usepackage{multirow,graphicx}
\usepackage{tabularx}
\usepackage{caption}

\usepackage[utf8]{inputenc}
\usepackage{color}

\setitemize{parsep=6pt,itemsep=0pt,leftmargin=*,labelsep=5.5mm}
\setenumerate{parsep=6pt,itemsep=0pt,leftmargin=*,labelsep=5.5mm}
\setlist[description]{itemsep=0mm}

 \usepackage[labelformat=simple]{subcaption}

\DeclareCaptionLabelFormat{subcaptionlabel}{\normalfont(\textbf{#2}\normalfont)}
\firstpage{1} 
\pubvolume{7}
\issuenum{0}\updates{yes}
\articlenumber{0}
\pubyear{2018}
\copyrightyear{2018}
\history{Received: 14 May 2018; Accepted: 26 May 2018; Published: date}

\Title{LiDAR and Camera Detection Fusion in a~Real-Time Industrial Multi-Sensor Collision Avoidance System}

\Author{{Pan Wei} *, Lucas Cagle, Tasmia Reza, John Ball and James Gafford}
\AuthorNames{Pan Wei, Lucas Cagle, Tasmia Reza, John Ball, Jim Gafford}

\address[1]{Center for Advanced Vehicular Systems (CAVS), Mississippi State University, Mississippi State, \mbox{MS 39759, USA}; ldc290@msstate.edu (L.C.); tr1044@msstate.edu (T.R.); jeball@ece.msstate.edu (J.B.); \mbox{gafford@cavs.msstate.edu (J.G.)}}

\corres{\hangafter=1 \hangindent=1.05em \hspace{-0.82em}Correspondence: pw541@msstate.edu; Tel.: +1-662-312-1916}

\abstract{Collision avoidance is a~critical task in many applications, such as ADAS (advanced driver-assistance systems), industrial automation and robotics. In an industrial automation setting, certain areas should be off limits to an automated vehicle for protection of people and high-valued~assets. These areas can be quarantined by mapping (e.g., GPS) or via beacons that delineate a~no-entry~area. We propose a~delineation method where the industrial vehicle utilizes a~LiDAR {(Light Detection and Ranging)} and a~single color camera to detect passive beacons and model-predictive control to stop the vehicle from entering a~restricted space. The beacons are standard orange traffic cones with a~highly reflective vertical pole attached. The LiDAR can readily detect these beacons, but suffers from false positives due to other reflective surfaces such as worker safety vests. Herein, we put forth a~method for reducing false positive detection from the LiDAR by projecting the beacons in the camera imagery via a~deep learning method and validating the detection using a~neural network-learned projection from the camera to the LiDAR space. Experimental data collected at Mississippi State University's Center for Advanced Vehicular Systems (CAVS) shows the effectiveness of the proposed system in keeping the true detection while mitigating false positives.}


\begin{document}

\section{Introduction}
\label{sec:Intro}
Collision avoidance systems are important for protecting people's lives and preventing property~damage. Herein, we present a~real-time industrial collision avoidance sensor system, which is designed to not run into obstacles or people and to protect high-valued equipment. The system utilizes a~scanning LiDAR and a~single RGB camera. A passive beacon is utilized to mark off quarantined areas where the industrial vehicle is not allowed to enter, thus preventing collisions with high-valued equipment. A front-guard processing mode prevents collisions with objects directly in front of the~vehicle.

To provide a~robust system, we utilize a~Quanergy eight-beam LiDAR and a~single RGB camera. The LiDAR is an active sensor, which can work regardless of the natural illumination. It can accurately localize objects via their 3D reflections. However, the LiDAR is monochromatic and cannot differentiate objects based on color. Furthermore, for objects that are far away, the LiDAR may only have one to two beams intersecting the object, making reliable detection problematic. Unlike LiDAR, RGB cameras can make detection decisions based on texture, shape and color. An RGB stereo camera can be used to detect objects and to estimate 3D positions. However, stereo cameras require extensive processing and often have difficulty estimating depth when objects lack textural cues. On the other hand, a~single RGB camera can be used to accurately localize objects in the image itself (e.g., determine bounding boxes and classify objects). However, the resulting localization projected into 3D space is poor compared to the LiDAR. Furthermore, the camera will degrade in foggy or rainy environments, whereas the LiDAR can still operate effectively. Herein, we utilize both the LiDAR and the RGB camera to accurately detect (e.g., identify) and localize objects. The focus of this paper is the fusion method, and we also highlight the LiDAR detection since it is not as straightforward as the camera-based detection. The~contributions of this paper are:
\begin{itemize}
  
  \item We propose a~fast and efficient method that learns the projection from the camera space to the LiDAR space and provides camera outputs in the form of LiDAR detection (distance and angle).
  
  \item We propose a~multi-sensor detection system that fuses both the camera and LiDAR detections to obtain more accurate and robust beacon detections.
  
  \item The proposed solution has been implemented using a~single Jetson TX2 board (dual CPUs and a~GPU) board to run the sensor processing and a~second TX2 for the model predictive control (MPC) system. The sensor processing runs in real time (5 Hz).
  
  \item The proposed fusion system has been built, integrated and tested using static and dynamic scenarios in a~relevant environment. Experimental results are presented to show the fusion~efficacy.

\end{itemize}
 
This paper is organized as follows: Section \ref{sec:background} introduces LiDAR detection, camera detection and~the fusion of LiDAR and camera. Section \ref{sec:system} discusses the proposed fusion system, and Section \ref{sec:experiment} gives an experimental example showing how fusion and the system work. In Section \ref{sec:results}, we compare the results with and without fusion. Finally, Section \ref{sec:conclusion} contains conclusions and future work.

\section{Background} \vspace{-6pt}
\label{sec:background}

\subsection{Camera Detection} 
Object detection from camera imagery involves both classification and localization of each object in which we have interest. We do not know ahead of time how many objects we expect to find in each image, which means that there is a~varying number of outputs for every input image. We also do not know where these objects may appear in the image, or what their sizes might be. As a~result, the object detection problem becomes quite challenging.

With the rise of deep learning (DL), object detection methods using DL \citep{redmon2016you, redmon2016yolo9000, ren2015faster, szegedy2014scalable, li2016r, liu2016ssd} {have} surpassed many traditional methods \citep{dalal2005histograms, felzenszwalb2010object} in both accuracy and speed. 
Based on the DL detection methods, there~are also systems \citep{wei2015measuring, wei2016multi, wei2018fusion} improving the detection results in a~computationally-intelligent way.
Broadly speaking, there are two approaches for image object detection using DL. One approach is based on region proposals. Faster R-CNN \citep{ren2015faster} is one example. This method first runs the entire input image through some convolutional layers to obtain a~feature map. Then, there is a~separate region proposal network, which uses these convolutional features to propose possible regions for detection. Last, the rest of the network gives classification to these proposed regions. As there are two parts in the network, one for predicting the bounding box and the other for classification, this kind of architecture may significantly slow down the processing speed. Another type of approach {uses} one network for both predicting potential regions and for label classification. One example is you only look once (YOLO)~\citep{redmon2016you,redmon2016yolo9000}. Given an input image, YOLO first divides the image into coarse grids. For each grid, there is a~set of base bounding boxes. For each base bounding box, YOLO predicts offsets off the true location, a~confidence score and classification scores if it thinks that there is an object in that grid location. YOLO is fast, but sometimes can fail to detect small objects in the image. 

\subsection{LiDAR Detection}
For LiDAR detection, the difficult part is classifying points based only on a~sparse 3D point cloud. One approach \citep{lin2014eigen} uses eigen-feature analysis of weighted covariance matrices with a~support vector machine (SVM) classifier. However, this method is targeted at dense airborne LiDAR point clouds. In another method \citep{golovinskiy2009shape}, the feature vectors are classified for each candidate object with respect to a~training set of manually-labeled object locations. With its rising popularity, DL has also been used for 3D object classification. Most DL-based 3D object classification problems involve two steps: deciding a~data representation to be used for the 3D object and training a~convolutional neural network (CNN) on that representation of the object. VoxNet \citep{maturana2015voxnet} is a~3D CNN architecture for efficient and accurate object detection from LiDAR and RGBD point clouds. An example of DL for volumetric shapes is the Princeton ModelNet dataset \citep{wu20153d}, which has proposed a~volumetric representation of the 3D model and a~3D volumetric CNN for classification. However, these solutions also depend on a~high density (high beam count) LiDAR, so they would not be suitable for a system with an eight-beam Quanergy M8.

\subsection{Camera and LiDAR Detection Fusion}

Different sensors for object detection have their advantages and disadvantages. Sensor fusion integrates different sensors for more accurate and robust detection. For instance, in object detection, cameras can provide rich texture-based and color-based information, which LiDAR generally lacks. On the other hand, LiDAR can work in low visibility, such as at night or in moderate fog or rain. \mbox{Furthermore, for the} detection of the object position relative to the sensor, the LiDAR can provide a~much more accurate spatial coordinate estimation compared to a~camera. As both camera and LiDAR have their advantages and disadvantages, when fusing them together, the ideal algorithm should fully utilize their advantages and eliminate their disadvantages.

One approach for camera and LiDAR fusion uses extrinsic calibration. 
Some works on this approach \citep{gong2013extrinsic, park2014calibration, garcia2013lidar} use various checkerboard patterns or look for corresponding points or edges in both the LiDAR and camera imagery. Other works on this approach \citep{levinson2013automatic, gong20133d, napier2013cross, pandey2015automatic, castorena2016autocalibration} look for corresponding points or edges in both the LiDAR and camera imagery in order to perform extrinsic calibration. However, for~the majority of works on this approach, the LiDARs they use have either 32 or 64 beams, which~provide relatively high vertical spatial resolution, but are prohibitively expensive. Another example is \citep{li20152d}, which estimates transformation matrices between the LiDAR and the camera. Other similar examples include \citep{zhang2004extrinsic, vasconcelos2012minimal}. However, these works are only suitable for modeling of indoor and short-range environments. 
Another approach uses a~similarity measure. One example is \citep{mastin2009automatic}, which automatically registers LiDAR and optical images. This approach also uses dense LiDAR measurements. A third kind of approach uses stereo cameras and LiDAR for fusion. One example is \citep{maddern2016real}, which fuses sparse 3D LiDAR and dense stereo image point clouds. However, the matching of corresponding points in stereo images is computationally complex and may not work well if there is little texture in the images. Both of these approaches require dense point clouds and would not be effective with the smaller LiDARs such as the Quanergy M8. Comparing with previous approaches, the proposed approach in this paper is unique in two aspects: (1) a~single camera and relatively inexpensive eight-beam LiDAR are used for an outdoor collision avoidance system; and (2) the proposed system is a~real-time system.

\subsection{Fuzzy Logic}
When Zadeh invented the term ``fuzzy'' to describe the ambiguities in the world, he developed a~language to describe his ideas. One of the important terms in the ``fuzzy'' approach is the fuzzy set, which is defined as a~class of objects with a~continuum of grades of membership \citep{zadeh1965fuzzy}. The relation between a linguistic variable and the fuzzy set is described as ``a linguistic variable can be interpreted using fuzzy sets'' in \citep{ross2010fuzzy}. This~statement basically means a~fuzzy set is the mathematical representation of linguistic variables. An~example of a fuzzy set is ``a class of tall students''. It is often denoted as $ \underset{^\sim}A$ or sometimes $A$. The~membership function assigns to each object a~grade of membership ranging between zero and one~\citep{zadeh1965fuzzy}.

Fuzzy logic is also proposed in \citep{zadeh1965fuzzy} by Zadeh. Instead of giving either true or false conclusions, fuzzy logic uses degrees of truth. The process of fuzzy logic includes three steps. First, input is fuzzified into the fuzzy membership function. Then, IF-THEN rules {are} applied to produce the fuzzy output function. Last, the fuzzy output function is de-fuzzified to get specific output values. \mbox{Thus, fuzzy logic} can be used to reason with inputs that have uncertainty.

Zhao et al. \citep{zhao2012fusion} applied fuzzy logic to fuse Velodyne 64-beam LiDAR and camera data. They~utilized LiDAR processing to classify objects and then fused the classifications based on the bounding box size (tiny, small, middle or large), the image classification result, the spatial and temporal context of the bounding box and the LiDAR height. Their system classified outputs as greenery, middle or obstacle. Again, the use of a~Velodyne 64-beam LiDAR is prohibitively expensive for our application.

Fuzzy logic provides a~mathematically well-founded mechanism for fusion with semantic meaning, such as ``the LiDAR detects an object with {high confidence}''. This semantic statement can be quantified mathematically by the fuzzy function. The fuzzy logic system can then be used to fuse the fuzzified inputs, using the rules of fuzzy logic. Herein, we apply fuzzy logic to combine confidence scores from the camera and the LiDAR to obtain a~detection score for the final fusion result.

%
%
\section{Proposed System} \vspace{-6pt}
\label{sec:system}

\subsection{Overview}

\begin{figure}[H]
\centering
\caption{Collision avoidance system block diagram. IMU = inertial measurement unit. \mbox{GPU = graphics processing} unit. CPU = central processing unit. SVM = support vector machine. \mbox{CNN = convolutional neural} network. Figure best viewed in color.}
\label{fig:SystemBlockDiagram}
\includegraphics[width=6in]{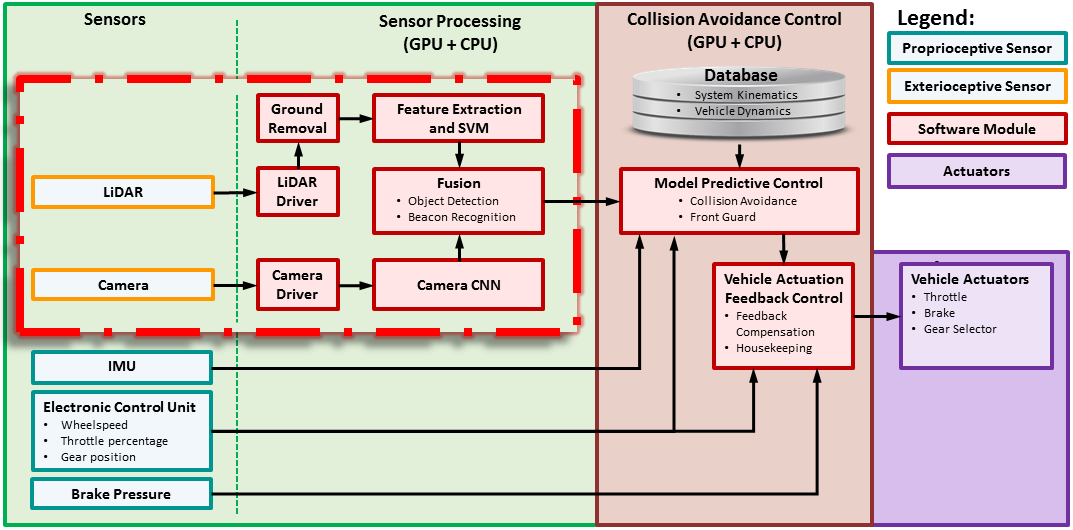}
\end{figure}

\begin{figure}[H]
\centering
\caption{Using passive beacons to delineate a~no-entry area.}
\label{fig:protect}
\includegraphics[width=4in]{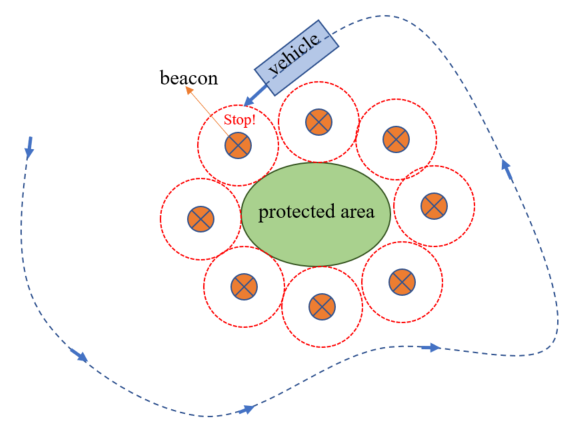}
\end{figure}

The proposed system architecture block diagram is shown in Figure \ref{fig:SystemBlockDiagram}. In the figure, the camera and LiDAR are exteroceptive sensors, and their outputs go to the signal processing boxes. The LiDAR signal processing is discussed in Section \ref{subsec:LiDARDetection}, and the camera signal processing is discussed in Section~\ref{subsec:CameraDetection}. The LiDAR detects beacons and rejects other objects. The LiDAR signal processing output gives beacon locations as a~distance in meters (in front of the industrial vehicle), the azimuth angle in degrees and~a~discriminant value, which is used to discriminate beacons from non-beacons. The camera reports detections as relative bounding box locations in the image, as well as the~confidence of the detection and classification. The LiDAR and camera information is fused to create more robust detections. The~fusion is discussed in Section \ref{subsec:LiDARCameraDecisionFusion}. Figure~\ref{fig:protect} shows one application of using this system for collision avoidance. In this figure, the green area in the middle is quarantined and designated as a~non-entry area for industrial vehicles.

%
%
\unskip

As shown in Figure \ref{fig:SystemBlockDiagram}, the vehicle also has proprioceptive sensors, including wheel speed, steering~angle, throttle position, engine control unit (ECU), two brake pressure sensors \mbox{and an inertial} measurement unit (IMU). These sensors allow the vehicle actuation and feedback control system to interpret the output of the system model MPC system and send commands to the ECU to control throttle and breaking. 

The MPC and vehicle actuation systems are critical to controlling the industrial vehicle, but are not the focus of this paper. These systems model the vehicle dynamics and system kinematics and~internally maintain a~virtual barrier {around the detected} objects. MPC by definition is a~model-based approach, which incorporates a~predictive control methodology to approximate optimal system control. Control solutions are determined through the evaluation of a~utility function comparing computations of a~branching network of future outcomes based on observations of the present environment \citep{Ersal2015,Alrifaee2017,Iagnemma2010}. The implemented MPC algorithm, to which this work is applied, predicts collision paths for the vehicle system with detected objects, such as passive beacons. Vehicle speed control solutions are computed for successive future time steps up to the prediction horizon. A~utility function determines optimal control solutions from an array of possible future control solutions. Optimal control solutions are defined in this application as solutions that allow for the operation of the vehicle at a~maximum possible speed defined as the speed limit, which ensures the vehicle is operating within the control limitations of the brake actuation subsystem such that a~breach of a~virtual boundary around detected objects may be prevented. {In the implemented system,} the prediction horizon may exceed 5 s. At the maximum possible vehicle speed, this horizon is consistent with the detection horizon of the sensor network. The development of the system model is partially described in \citep{liu2017development,Davenport2018}. 

\begin{figure}[H]
\centering
\caption{Proposed fusion system high-level block diagram.}
\label{fig:proposedsystem}
\includegraphics[width=6in]{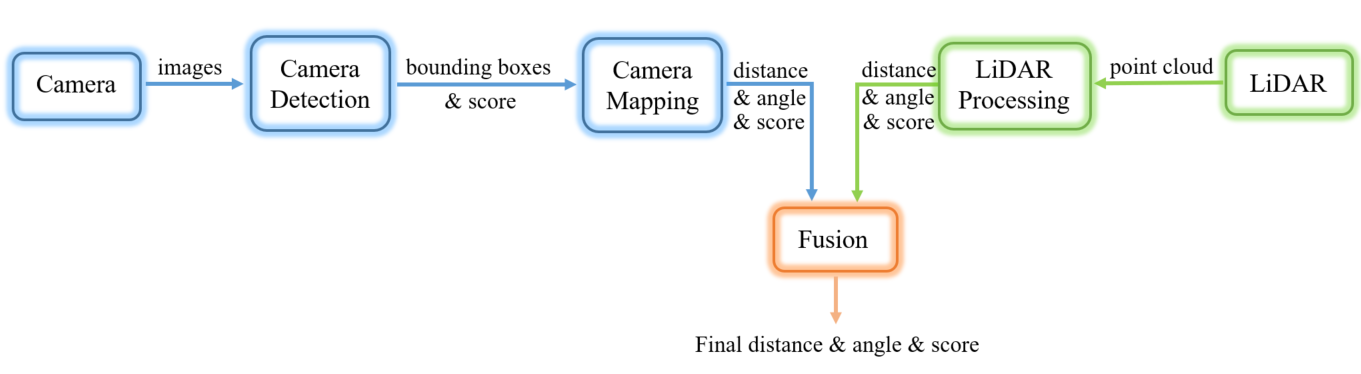}
\end{figure}

The proposed fusion system is shown in Figure \ref{fig:proposedsystem}. In this system, we first obtain images from the camera, then through the camera object detection system, we estimate the bounding box coordinates of beacons together with the confidence scores for each bounding box. Next, the detection bounding box from the single camera is mapped through a~neural network (NN) into a~distance and angle estimate, in order to be able to fuse that information with the distance and angle estimates of the LiDAR processing.

On the other side, we have point cloud information from the LiDAR, then through LiDAR data processing, we estimate the distance and angle information of the detected beacons by the LiDAR. Through the LiDAR processing, we also obtain a~pseudo-confidence score. With the LiDAR detection results in the form of distance and angle from both the camera and LiDAR, we fuse them together to obtain a~final distance, angle and confidence score for each detection. We give more details about this in Section \ref{subsec:detectionfusion}. 

In our application, the maximum speed of the vehicle is about 5 m per second. In order to have enough reaction time for collision avoidance, the detection frequency needs to be 5 Hz or above. We~accomplish this requirement for sensor processing on one NVIDIA Jetson TX2 and use another TX2 for the MPC controller. More details are given in Section \ref{subsec:realtime}.

%
%

\subsection{Real-Time Implementation} \label{subsec:realtime}

To achieve the desired 5-Hz update rate for real-time performance, multiple compromises were made to improve software speed enough to meet the requirements. Notably, a~naive approach for removal of ground points from the LiDAR point cloud was used instead of a~more complex, and~accurate, method. Furthermore, a~linear SVM, with a~limited feature set, was used as opposed to a~kernel SVM or other non-linear methods. The specifics of these methods are covered in Section \ref{subsec:LiDARDetection}.

To manage intra-process communication between sensor drivers and signal processing functions, the Robot Operating System (ROS) was used \citep{Quigley09}. ROS provides already-implemented sensor drivers for extracting point cloud and image data from the LiDAR and camera, respectively. Additionally, ROS~abstracts and simplifies intra-process data transfer by using virtualized Transmission Control Protocol (TCP) connections between the individual ROS processes, called nodes. As a~consequence of this method, multiple excess memory copies must be made, which begin to degrade performance with large objects like LiDAR point clouds. To ameliorate this effect, ROS Nodelets \citep{wiki2018nodelet} were used for appropriate LiDAR processes. Nodelets builds on ROS Nodes to provide a~seamless interface for pooling multiple ROS processes into a~single, multi-threaded process. This enables efficient zero-copy data transfers between signal processing steps for the high bandwidth LiDAR computations. This~proved essential due to the limited resources of the TX2, with central processing unit (CPU) utilization averaging at over 95\% even after Nodelets optimization.

\begin{figure}[H]
\centering
\caption{ROS dataflow on Jetson TX2.}
\label{fig:roshw}
\includegraphics[width=3.5in]{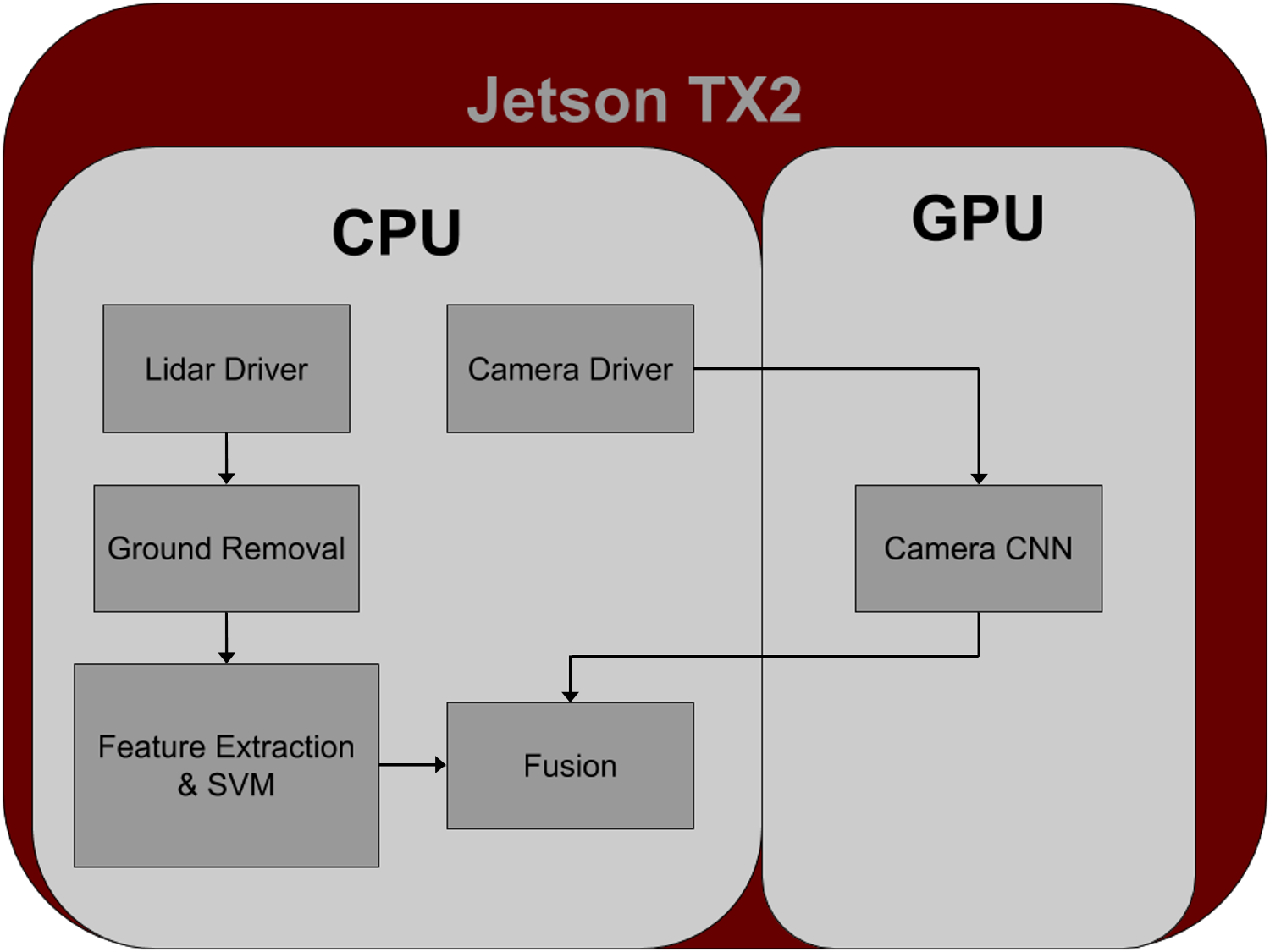}
\end{figure}

In Figure \ref{fig:roshw}, the high-level flowchart of the relevant ROS nodes can be seen along with their execution location on the Jetson TX2's hardware. The NVIDIA Jetson TX2 has a~256 CUDA {(Compute Unified Device Architecture)} core GPU, a~Dual CPU, 8 GB of memory and supports various interfaces including Ethernet \citep{NVIDIA_TX2}. As the most computationally-intensive operation, the camera CNN has the entire graphical processing unit (GPU) allocated to it. The sensor drivers, signal processing and fusion nodes are then relegated to the CPU. Due to the high core utilization from the LiDAR, care has to be taken that enough CPU resources are allocated to feed images to the GPU or significant throttling of the CNN can occur.

%
%

\subsection{Beacon}

\begin{figure}[H]
\centering
\caption{Beacon.}
\label{fig:beacon}
\includegraphics[width=3.5in]{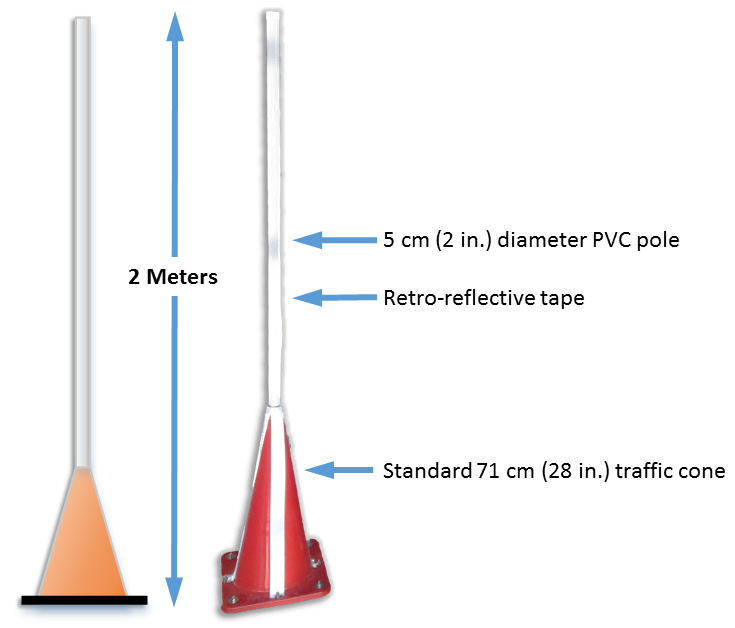}
\end{figure}

The beacon is constructed of a~high-visibility 28'' orange traffic cone with a~2'' diameter highly-reflective pole extending two meters vertically. A beacon is shown in Figure \ref{fig:beacon}. The beacon presents as a~series of high intensity values in a~LiDAR scan and thus provides a~high signal level compared to most background objects. A beacon delineates an area in the industrial complex that is off limits to the vehicle, usually to protect high-valued assets.

%
%

\subsection{Camera Detection}
\label{subsec:CameraDetection}

Herein, we apply a~deep-learning model called YOLO \citep{redmon2016you, redmon2016yolo9000} to detect objects present in the single camera image. YOLO outperforms most other deep learning methods in speed, which is crucial for our application. YOLO achieves such speed by combining both region and class prediction into one~network, unlike region-based methods \citep{li2016r, ren2015faster, girshick2015fast}, which have a separate network for region proposals. 

\begin{figure}[H]
\centering
\caption{Labeling for camera detection.}
\label{fig:labeling}
\includegraphics[width=4.5in]{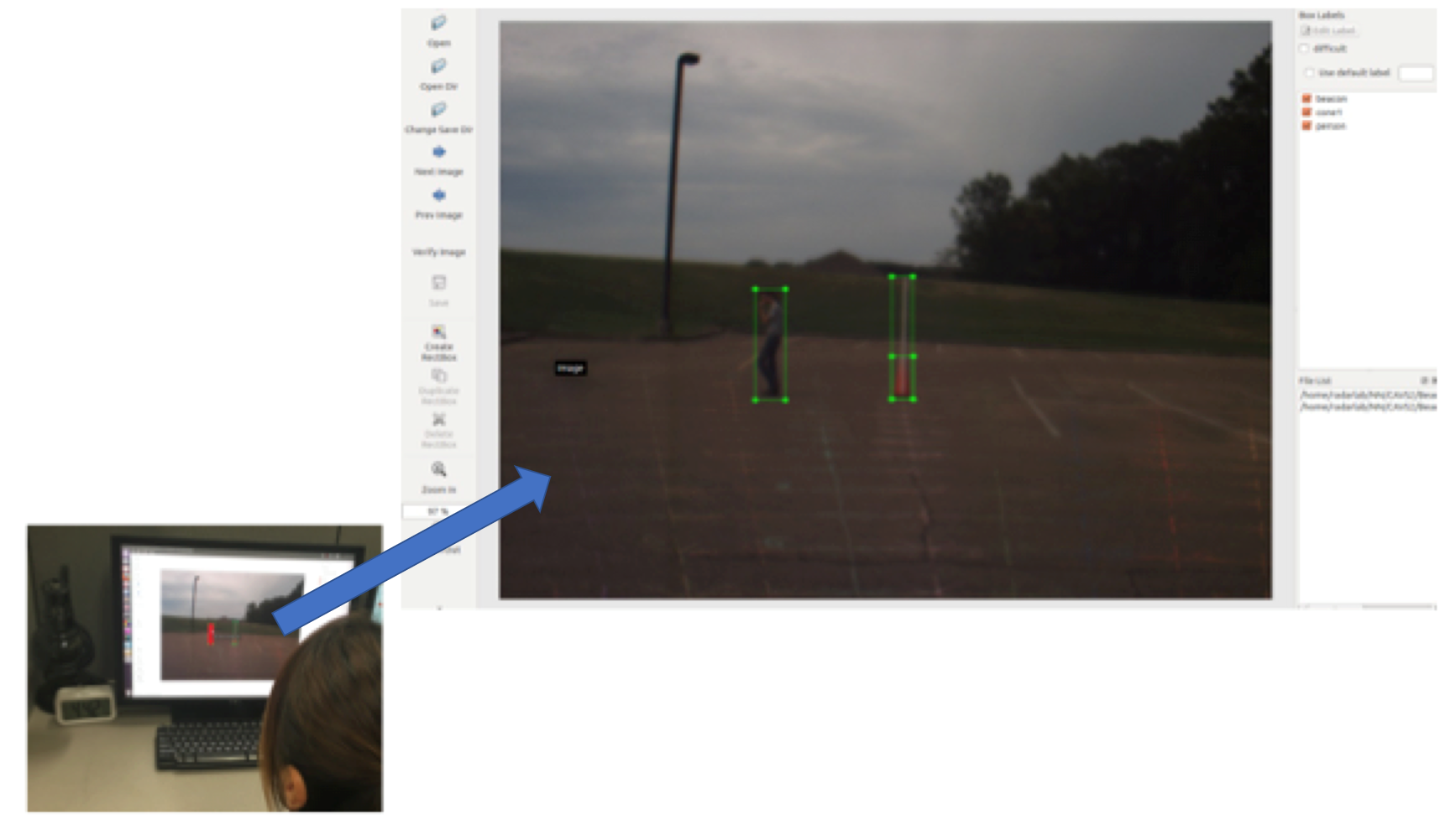}
\end{figure}

A large training dataset of beacon images {presented in Section} \ref{sec:experiment} was collected on different days, {different times of day} and weather conditions {(e.g., full sun, overcast, etc.)} and was hand-labeled by our team members. Figure~\ref{fig:labeling} shows how we draw bounding boxes and assign labels on an image from~the camera. 

\begin{figure}[H]
\centering
\begin{subfigure}[b]{0.45\textwidth}
\includegraphics[width=\textwidth]{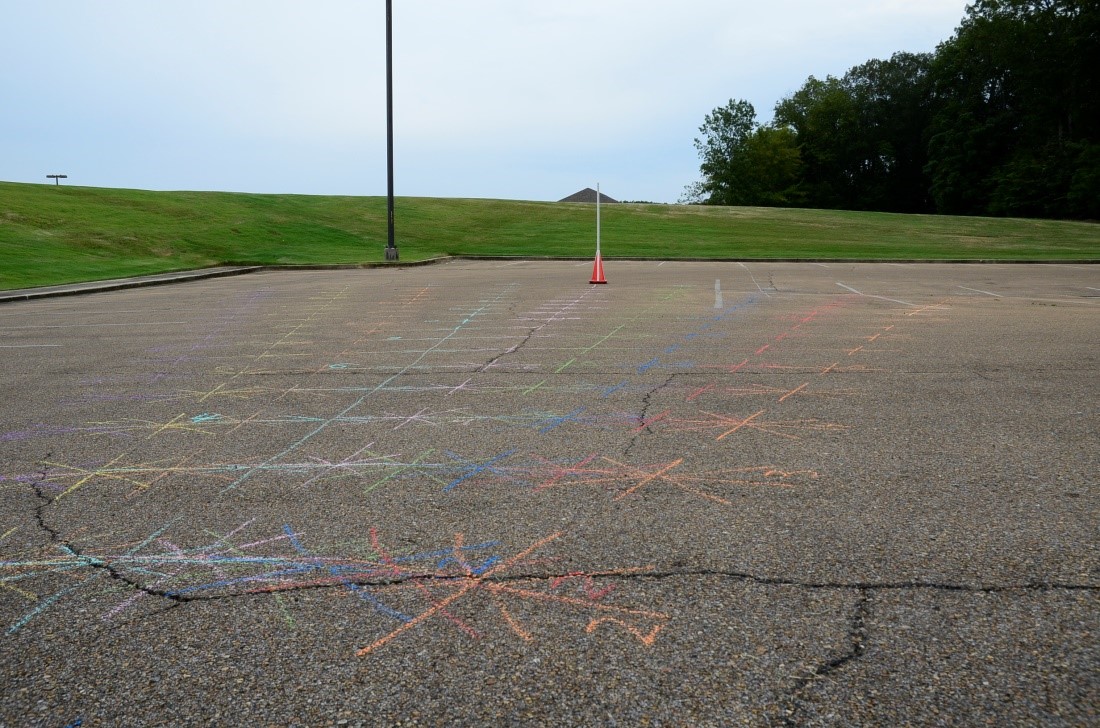}
\caption{}
\label{subfig:groundmarking1}
\end{subfigure}
\begin{subfigure}[b]{0.43\textwidth}
\includegraphics[width=\textwidth]{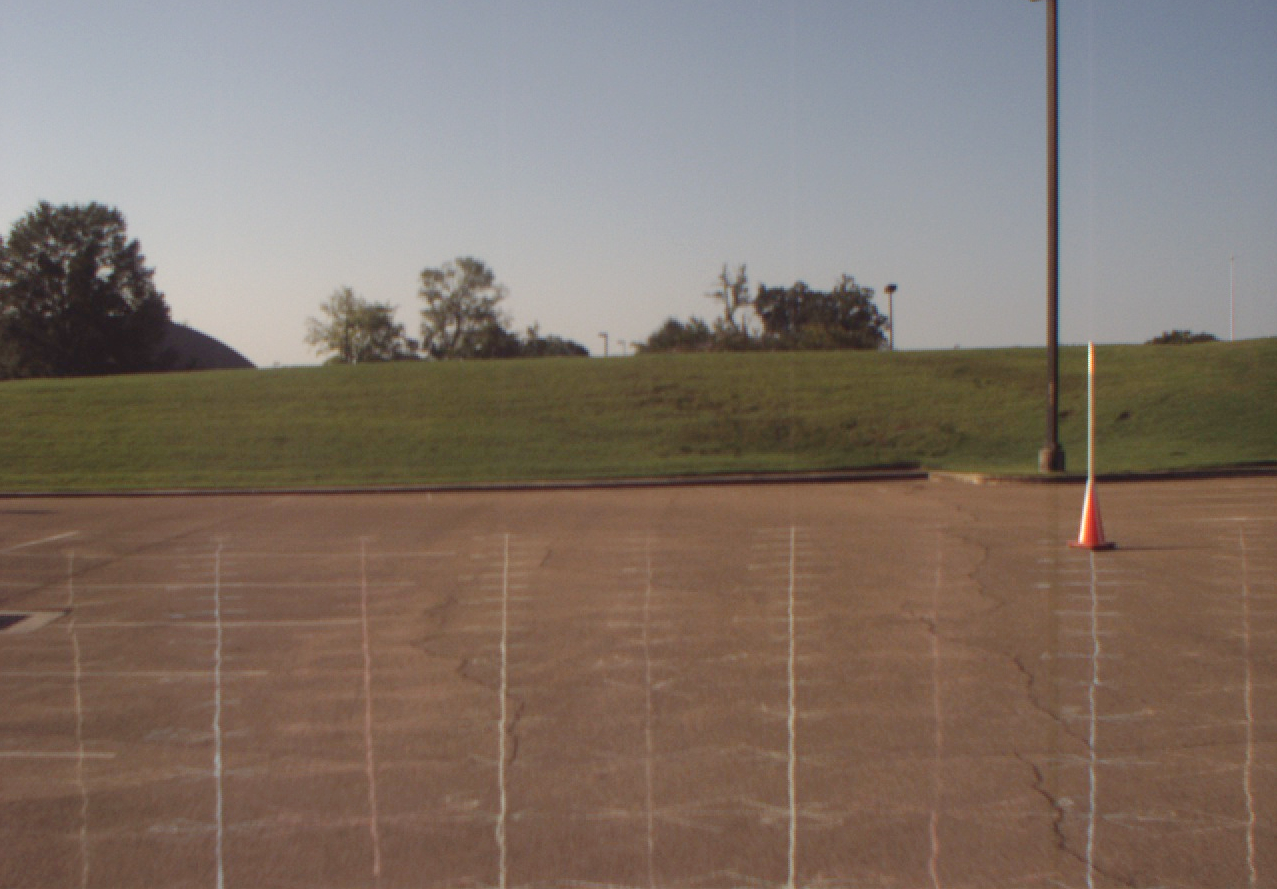}
\caption{}
\label{subfig:groundmarking2}
\end{subfigure}
\caption{(\textbf{a}) Markings on the ground. (\textbf{b}) View from the camera. Markings on the ground covering the full FOV and detection range of the camera.}
\label{fig:groundmarking}
\end{figure}

The dataset we collected covers the camera's full field of view (FOV) ($-20^{\circ}$ to $20 ^{\circ}$ {azimuth}) and~detection range (5 to 40 m) (FLIR Chameleon3 USB camera with Fujinon 6-mm lens). It reliably represents the detection accuracy from different angles and distances. Figure~\ref{fig:groundmarking} shows the markings on the ground that help us to locate the beacon at different angles and distances. These tests were performed early in the project to allow us to estimate the LiDAR detection accuracies.

\begin{figure}[H]
\centering
\caption{Camera detection example. Best viewed in color.}
\label{fig:camera}
\includegraphics[width=3.5in]{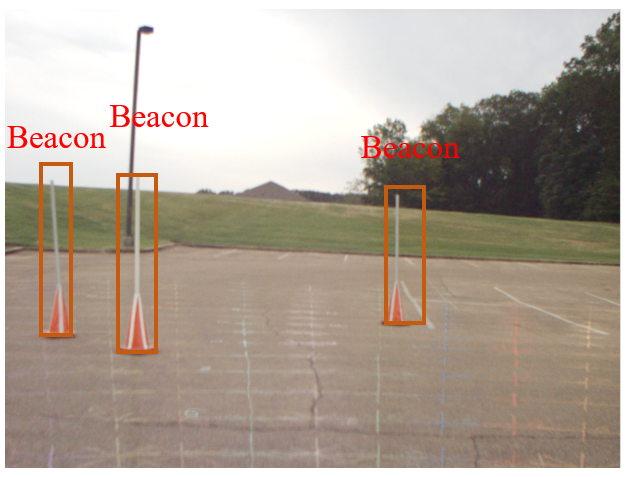}
\end{figure}

All of the beacons have a~highly consistent construction, with near-identical size, shape~and~coloring. These characteristics allow the detection algorithm to reliably recognize beacons at a~range of up to 40 m under a~variety of lighting conditions. Figure \ref{fig:camera} shows one example of the camera detection of beacons.

%
%

\subsection{LiDAR Front Guard}
\label{subsec:LiDARFrontGuard}
The LiDAR detection mode is designed to detect beacons. However, other objects can be in the direct or near-direct path of the industrial vehicle. A front guard system is implemented with the LiDAR. Since many objects lack any bright points, thus being missed by the LiDAR beacon detection, we also employed a~front-guard system, which clusters points in a~rectangular area above the ground directly in front of the industrial vehicle. The front guard is then used to keep the vehicle from striking obstacles or people. After ground point removal, any remaining points within a~rectangular solid region directly in front of the vehicle will be detected and reported to the control system.

\subsection{LiDAR Beacon Detection}
\label{subsec:LiDARDetection}
This section discusses the LiDAR's beacon detection system. 

\subsubsection{Overview}

\begin{figure}[H]
\centering
\caption{LiDAR beam spacing visualization and beam numbering. The LiDAR is the small black box on the left. Beam 6 is the horizontal beam (assuming the LiDAR is level).}
\label{fig:lidar_rings}
\includegraphics[width=5.0in]{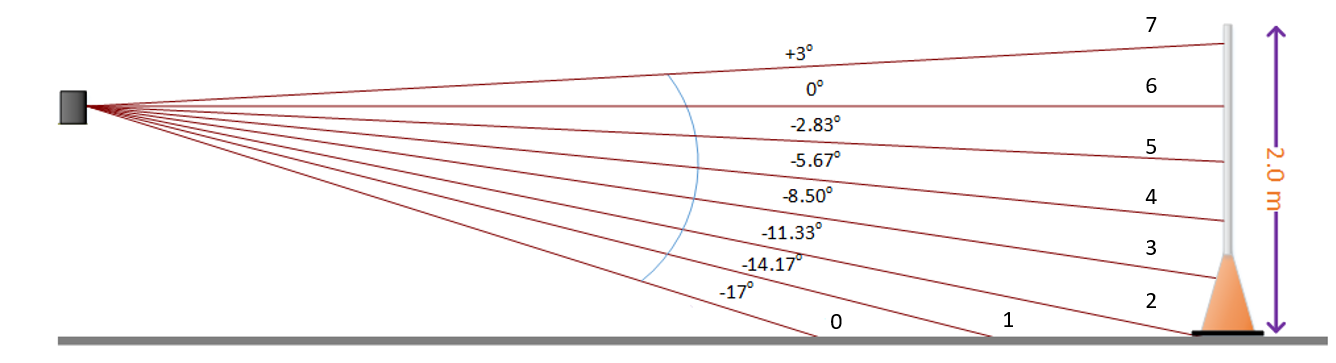}
\end{figure}

\begin{figure}[H]
\centering
\caption{Detection via LiDAR.}
\label{fig:lidar}
\includegraphics[width=6in]{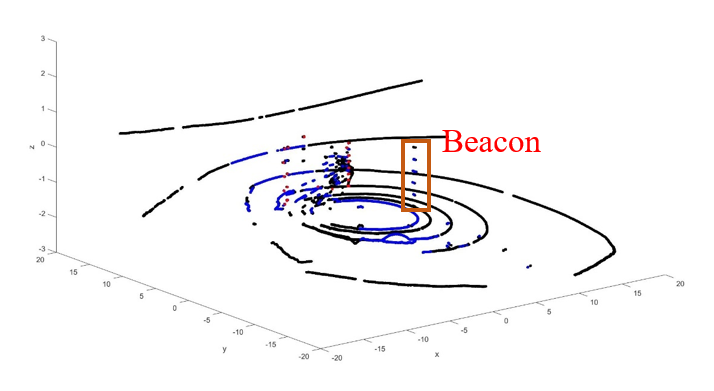}
\end{figure}

The LiDAR data we collect {are} in the form of a~three-dimensional (3D) and 360 degree field-of-view (FOV) point cloud, which consists of $x$, $y$, $z$ coordinates, along with intensity and beam number (sometimes called ring number). The coordinates $x$, $y$ and $z$ represent the position of each point relative to the origin centered within the LiDAR. Intensity represents the strength of returns and is an~integer for this model of LiDAR. Highly reflective objects such as metal or retro-reflective tape will have higher intensity values. {The beam number represents in which beam the returned point is located.} The LiDAR we use transmits eight beams. The beam numbers are shown in Figure \ref{fig:lidar_rings}. Assuming the LiDAR is mounted level to the ground, then Beam 7 is the top beam and is aimed upward approximately~$3^{\circ}$. Beam 6 points horizontally. The beams are spaced approximately $3^{\circ}$ apart. Figure \ref{fig:lidar} shows one scan of the LiDAR point cloud. In this figure, the ground points are shown with black~points. The beacon presents as a~series of closely-spaced points in the horizontal direction, as~shown in the square.

{General} object detection in LiDAR is a~difficult and unsolved problem in the computer vision community. Objects appear at different scales based on their distance to the LiDAR, and they can be viewed from any angle. However, in our case, we are really only interested in detecting the beacons, which all have the same geometries. In our application, the industrial vehicle can only go so fast, so~that we only need to detect beacons about 20 m away or closer. The beacons are designed to have very bright returns. However, other objects can also have bright returns in the scene, such~as people wearing safety vests with retro-reflective tape, or other industrial vehicles. Herein, we~propose a~system that first identifies clusters of points in the scene that present with bright returns. Next,~a~linear SVM classifier is used to differentiate the beacon clusters from non-beacon clusters.

%
%
\unskip

%
%

The LiDAR detection system works by utilizing intensity-based and density-based clustering (a modified DBSCAN algorithm) on the LiDAR point cloud. The system then examines all points near the cluster centroid by extracting features and using a~linear SVM to discriminate beacons from non-beacons. Details of these algorithm are discussed in the subsections below.

\subsubsection{LiDAR Clustering}
A modified DBSCAN clustering algorithm \citep{ester1996density}, which clusters based on point cloud density, as well as intensity is used to cluster the bright points, as shown in Algorithm \ref{alg:LiDAR_bright_clustering}. The cluster parameter $\epsilon$ was empirically determined to be 0.5 m based on the size of the beacon. Values much larger could group two nearby beacons (or a~beacon and another nearby object) together into one cluster, and we wanted to keep them as separate clusters.

In Equation (\ref{alg:LiDAR_bright_clustering}), the distances are estimated using Euclidean distances with only the $x$ (front-to-back) and $y$ (left-to-right) coordinates. Effectively, this algorithm clusters bright LiDAR points by projecting them down onto the
$x$-$y$ plane. Alternately, all three coordinates, $x,y,z$ could be used, but the features don not require this because they will be able to separate tall and short~objects. This approach is also more computationally efficient than using all three coordinates in the clustering~algorithm.

\vspace{12pt}
%
%
\begin{algorithm}[H]
\textbf{\\}
\KwIn{LiDAR point cloud $P = \{x_{j}, y_{j}, z_{j}, i_{j}, r_{j}\}$ with $NP$ points.}
\KwIn{High-intensity threshold: $T_{H}$.}
\KwIn{Cluster distance threshold: $\epsilon$ (meters).}
\KwIn{Ground $Z$ threshold: $T_{G}$ (meters).}

\textbf{\\}
\KwOut{Cluster number for each bright point in the point cloud.}

\textbf{\\Remove non-return points:\\}
\For{each point in the point cloud}
{
  Remove all non-return points (NaN's) from the the point cloud.
}

\textbf{\\Remove ground points:\\}
\For{each point in the modified point cloud}
{
  Remove all points with $Z$-values below $T_{G}$ from the point cloud.
}

\textbf{\\Remove non-bright points:\\}
\For{each point in the modified point cloud}
{
  Remove all points with intensity values below $T_{H}$ from the point cloud.
}

\textbf{\\Cluster bright points:\\}

Assign all points to Cluster 0.\\
Set $cl \leftarrow 0$.

\For{each point $P_{j}$ in the modified point cloud}
{
 \If{The point does not belong to a~cluster}
  {
   \textbf{Add point to cluster}\\
   Increment the number of clusters: $cl \leftarrow cl + 1$.\\
   Assign $P_{j}$ to cluster $cl$.\\
   Set the centroid of cluster $cl$ to $P_{j}$.\\
  
   \textbf{Scan through all remaining points and re-cluster if necessary.}\\
   \For{each point $P_{m}$ with $j < m \le NP$}
   {
     \If{Distance from point $P_{m}$ to the centroid of cluster $cl$ has distance $< \epsilon$}
     {
       Add $P_{m}$ to cluster $cl$.\\
       Recalculate the centroid of cluster $cl$.\\
     }   
   }

  }
}
\caption{{LiDAR bright pixel clustering.}} 
\label{alg:LiDAR_bright_clustering}
\end{algorithm}

\subsubsection{LiDAR Feature Extraction}
\label{subsec:LidarFeatureExtraction}
To extract the features, the ground points must be removed before feature processing occurs, or~else false alarms can occur. Since this industrial application has a~smooth, flat area for the ground, we~employ a~simple vertical threshold to remove ground points. If this system were to be generalized to areas with more varying ground conditions, ground estimation methods such as those in \citep{wang2016lidar, meng2010ground, rummelhard2017ground, rashidi2017ground, chang2008automatic, mikadlicki2017ground} could be utilized. However, for this particular application, this was not necessary. 

Next, the point intensities are compared to an empirically-determined threshold. The beacon is designed so that it provides bright returns to the LiDAR via the retro-reflective vertical pole. This~works~well, but there are also other objects in the scene that can have high returns, such~as other industrial vehicles with retro-reflective markings or workers wearing safety vests with retro-reflective~stripes. In order to classify objects as beacons and non-beacons, hand-crafted features are utilized (these are discussed {below}). After ground-removal and thresholding the intensity points, we are left with a~set of bright points. A second set of intensity points is also analyzed, which consists of all of the non-ground points (e.g., the point cloud after ground removal). The points in the bright point cloud are clustered. Beacons appear as tall, thin objects, whereas all other objects are either not as tall, or wider. We extract features around the cluster center in a~small rectangular volume centered at each object's centroid. We also extract features using a~larger rectangular volume also centered around the objects centroid. Features include counting the number of bright points in each region, determining the $x$, $y$ and $z$ extents of the points in each region, etc. Beacons mainly have larger values in the smaller region, while~other objects have values in the larger regions. 

\begin{figure}[H]
\centering
\caption{LiDAR detection regions (inner and outer) visualized from a~top-down view.}
\label{fig:DetectionRegionTopView}
\includegraphics[width=4.5in]{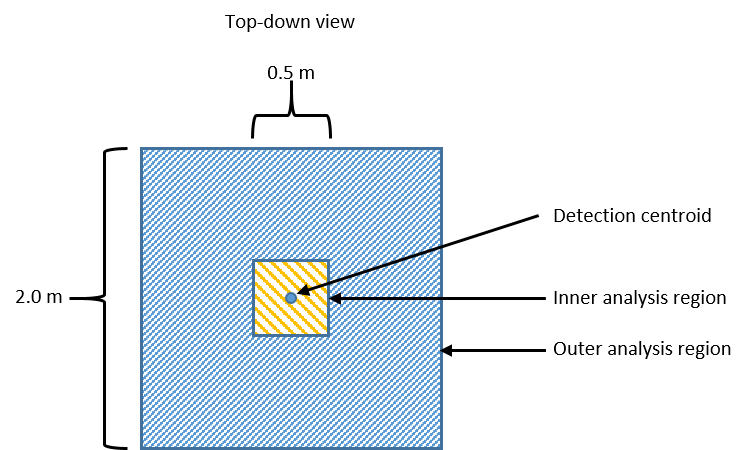}
\end{figure}

The two regions are shown in Figure \ref{fig:DetectionRegionTopView}. The idea of using an inner and an outer analysis region is that a~beacon will mostly have bright points located in the inner analysis region, while other objects, such as humans, other industrial vehicles, etc., will extend into the outer regions. \mbox{Equations (\ref{eqn:InnerRegion}) and (\ref{eqn:OuterRegion})} define whether a~LiDAR point $p_{j}$ with coordinates $\left( x_{j}, y_{j}, z_{j} \right)$ is in the inner region or outer region, respectively, where the object's centroid has coordinates $\left( x_{C}, y_{C}, z_{C} \right)$. Reference Figure \ref{fig:DetectionRegionTopView} for a~top-down illustration of the inner and outer regions. 

Figure \ref{fig:DetectionRegionTopView} shows an example beacon return with the analysis windows superimposed. Both the inner and outer analysis regions have $x$ and $y$ coordinates centered at the centroid location. The inner analysis region has a depth ($x$ coordinate) of 0.5 m and a width ($y$ coordinate) of 0.5 m, and the height includes all points with $z$ coordinate values of $-$1.18 m and above. The outer region extends 2.0 m in both $x$ and $y$ directions and has the same height restriction as the inner region. These values were determined based on the dimensions of the beacon and based on the LiDAR height. The parameters $\Delta x_{I}$, $\Delta y_{I}$ and $z_{MIN}$ define the inner region relative to the centroid coordinates. Similarly, the parameters $\Delta x_{O}$, $\Delta y_{O}$ and $z_{MIN}$ define the outer region relative to the centroid coordinates. A point is in the inner region if:
%
%
\begin{equation}
\label{eqn:InnerRegion}
 \begin{split}
   \left(x_{C} - \frac{\Delta x_{I}}{2} \right) &\le x_{j} \le \left(x_{C} + \frac{\Delta x_{I}}{2} \right) and \\
   \left(y_{C} - \frac{\Delta y_{I}}{2} \right) &\le y_{j} \le \left(y_{C} + \frac{\Delta y_{I}}{2} \right) and \\
   z_{MIN} &\ge z_{j} ,
 \end{split}
\end{equation}
 and a~point is in the outer region if:
\begin{equation}
\label{eqn:OuterRegion}
 \begin{split}
   \left(x_{C} - \frac{\Delta x_{O}}{2} \right) &\le x_{j} \le \left(x_{C} + \frac{\Delta x_{O}}{2} \right) and \\
   \left(y_{C} - \frac{\Delta y_{O}}{2} \right) &\le y_{j} \le \left(y_{C} + \frac{\Delta y_{O}}{2} \right) and \\
   z_{MIN} &\ge z_{j}.
 \end{split}
\end{equation}

%
%

In order to be robust, we also extract features on the intensity returns. A linear SVM was trained on a~large number of beacon and non-beacon objects, and each feature was sorted based on its ability to distinguish beacons from non-beacons. In general, there is a~marked increase in performance, then the curve levels out. It was found that ten features were required to achieve very high detection rates and low false alarms. These ten features were utilized in real time, and the system operates in real time at the frame rate of the LiDAR, 5 Hz. The system was validated by first running on another large test set independent of the training set, with excellent performance as a~result as presented in Section~\ref{subsubsubsec:svmlidar}. Then, extensive field tests were used to further validate the results {as presented in Section} \ref{sec:experiment}.

After detection, objects are then classified as beacons or non-beacons. These are appended to two separate lists and reported by their distance in meters from the front of the industrial vehicle, their~azimuth angle in degrees and their discriminate value. Through extensive experimentation, we~can reliably see beacons in the LiDAR's FOV from 3 to 20 m.

LiDAR feature extraction follows the overall algorithm shown in Algorithm \ref{alg:LiDAR_data_preproc}. The input is the LiDAR point cloud $P = \{x_{j}, y_{j}, z_{j}, i_{j}, r_{j} \}$, where $j$ is the index variable for the $j$-th point, and $x$, $y$, $z$, $i$ and $r$ refer to the $x$ point in meters, the $y$ point in meters, the $z$ point in meters, the intensity and the beam number, respectively, for the $j$-th point. Note that all coordinates are relative to the center of the LiDAR at point $(0,0,0)$. The $x$ coordinate is positive in front of the LiDAR, and negative behind. The $y$ coordinate is positive to the left of the LiDAR and negative to the right. The $z$ coordinate is positive above the LiDAR and negative below.

\vspace{12pt}

%
%
\begin{algorithm}[H]
\textbf{\\}
\KwIn{LiDAR point cloud $P = \{x_{j}, y_{j}, z_{j}, i_{j}, r_{j} \}$.}
\KwIn{Low-intensity threshold: $T_{L}$.}
\KwIn{High-intensity threshold: $T_{H}$.}
\KwIn{Ground $Z$ threshold: $T_{G}$ (meters).}

\textbf{\\}
\KwOut{Feature vector $\textbf{f}$.}

\textbf{\\Remove non-return points:\\}
\For{each point in the point cloud}
{
  Remove all non-return points (NaN's) from the point cloud.
}

\textbf{\\Remove ground points:\\}
\For{each point in the modified point cloud}
{
  Remove all points with $Z$-values below $T_{G}$ from the point cloud.
}

\textbf{\\Create threshold point clouds:\\}
Set $P_{HT} = \varnothing$. \\
Set $P_{LT} = \varnothing$. \\ 
\For{each point $P_{j}$ in the modified point cloud}
{
 
  \If{Point $P_{j}$ has intensity $\ge T_{H}$}
  {
   Add $P_{j}$ to $P_{HT}$.
  }
  \If{Point $P_{j}$ has intensity $\ge T_{L}$}
  {
   Add $P_{j}$ to $P_{LT}$.
  }
}

\textbf{\\Extract features:\\}
\text{Extract features $\textbf{f}$ using Algorithm \ref{alg:LiDAR_feature_ext}.}
\caption{LiDAR high-level feature extraction preprocessing.} 
\label{alg:LiDAR_data_preproc}
\end{algorithm}

\vspace{8pt}
In order to extract features for objects, any points that did not provide a~return to the LiDAR are removed. These points will present as NaN's (not a~number) in the LiDAR point cloud. \mbox{Next, the estimated} ground points are removed. 
The non-ground points are divided into two data subsets: The high-threshold (HT) data and the low-threshold (LT) data. The HT data points only contain high-intensity returns, while the LT data contain points greater than or equal to the low-intensity threshold. Both data subsets do not contain any ground points. The high-intensity threshold was set to 15 and the low-intensity threshold to zero. These threshold values were determined experimentally based on the examination of multiple beacon returns at various distances. \\

%
%
\begin{algorithm}[H]
\textbf{\\}
\KwIn{LiDAR high-intensity point cloud $P_{HT} = \{x_{j}, y_{j}, z_{j}, i_{j}, r_{j} \}$.}
\KwIn{LiDAR low-intensity point cloud $P_{LT} = \{x_{j}, y_{j}, z_{j}, i_{j}, r_{j} \}$.}
\KwIn{Inner region $x$-extent: $\Delta x_{I}$ (meters).}
\KwIn{Inner region $y$-extent: $\Delta y_{I}$ (meters).}
\KwIn{Outer region $x$-extent: $\Delta x_{O}$ (meters).}
\KwIn{Outer region $y$-extent: $\Delta y_{O}$ (meters).}
\KwIn{LiDAR height above ground: $Z_{L} = 1.4$ (meters).}

\textbf{\\}
\KwOut{Feature vector $\textbf{f}$.}

\textbf{\\Cluster the high-intensity point cloud:\\}
\For{each point in $\textbf{p}$ in the high-intensity point cloud}
{
 Cluster points and determine cluster centers.
}

\textbf{\\Calculate features:\\}
\For{each cluster center point $c = \left( x_{C},y_{C},z_{C} \right)$ in the point cloud}
{
 Determine all points in $P_{HT}$ in the inner region using Equation (\ref{eqn:InnerRegion}), and calculate Feature{s} 1, { 13 and 17 } from Table \ref{table:Feature_Descriptions}.\\
 Determine all points in $P_{HT}$ in the outer region using Equation (\ref{eqn:OuterRegion}), and calculate Feature 4 from Table \ref{table:Feature_Descriptions}.\\
 Determine all points in $P_{LT}$ in the inner region using Equation (\ref{eqn:InnerRegion}), and calculate Features 6, 7, 9 {10, 11, 14, 16 and 18} from Table \ref{table:Feature_Descriptions}.\\ 
 Determine all points in $P_{LT}$ in the outer region using Equation (\ref{eqn:OuterRegion}), and calculate Features 2, 3, 5 {8, 12, 15, 19 and 20} from Table \ref{table:Feature_Descriptions}.\\ 
}
\textbf{\\}
\text{Return $\textbf{f} = \left[f_{1},f_{2},f_{3},\cdots,f_{20} \right]$.}
\caption{{LiDAR feature extraction.}}
\label{alg:LiDAR_feature_ext}
\end{algorithm}

\vspace{8pt}

\begin{figure}[H]
\centering
\caption{Score value versus number of concatenated features.}
\label{fig:Score_vs_Num_features}
\includegraphics[width=5in]{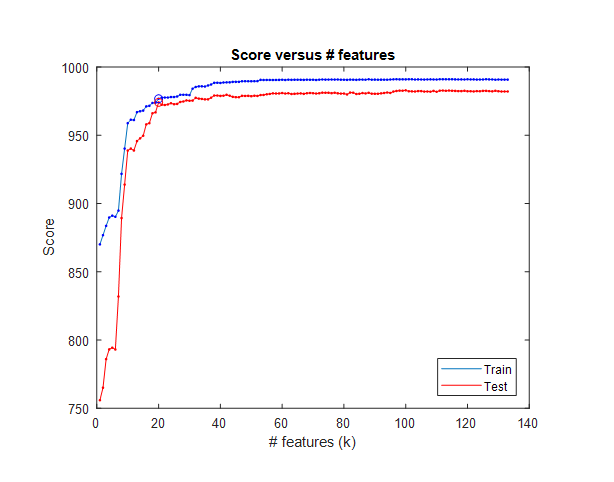}
\end{figure}

Table \ref{table:Feature_Descriptions} describes the extracted features. For example, Feature 1 is generated from the high threshold data, which consists solely of high-intensity points. The data is analyzed only for high-intensity points in the inner analysis region. This {feature} simply computes the $Z$ (height) extent, that is, the maximum $Z$ value minus the minimum $Z$ value. Note that some of the features are only processed on certain beams, such as Feature 2. The features are listed in order of their discriminate power in descending order, e.g., Feature 1 is the most discriminate, Feature 2 the next most discriminate, etc. {Initially, it was unknown how effective the features would be. We started with 134 features, and~these were ranked in order of their discriminating power using the score defined as:}
\begin{equation}
  SCORE = 500\frac{TP}{TP + FN} + 500\frac{TN}{TN + FP}
\end{equation}
where $TP$, $TN$, $FP$ and $FN$ {are the number of true positives, true negatives, false positive and false negatives, respectively } \citep{lillywhite2013feature}. {A higher score is better, and scores range from zero to 1000. This score was used since training with overall accuracy tended to highly favor one class and provided poor generalizability, which may be a~peculiarity of the chosen features and the training set. Figure} \ref{fig:Score_vs_Num_features} {shows the score values versus the number of features (where the features are sorted in descending score~order. The operating point is shown for} $M=20$ {features (circles in the plot). The features generalize fairly well since the testing curve is similar to the training curve. In our system, using 20 features provided a~good balance of performance versus computational complexity to compute the features. Each LiDAR scan (5-Hz scan rate) requires calculating features for each cluster. Experimentation showed that a total of 20~features was near the upper limit of what the processor could calculate and not degrade the LiDAR processing frame rate. }

%
%
\begin{table}[H]
\centering
\caption{Feature descriptions. Data subsets: HT = high threshold data. LT = low threshold data. The~LiDAR rings are shown in Figure \ref{fig:lidar_rings}.}
\label{table:Feature_Descriptions}
\begin{tabular}{cccl}
\toprule
\textbf{\begin{tabular}[c]{@{}c@{}}Feature\\ Number\end{tabular}} & \textbf{\begin{tabular}[c]{@{}c@{}}Data\\ Subset\end{tabular}} & \textbf{\begin{tabular}[c]{@{}c@{}}Analysis\\ Region\end{tabular}} & \multicolumn{1}{c}{\textbf{Description}} \\ \midrule
1 & HT & Inner & Extent of $Z$ in cluster. Extent \{$Z$\} = max\{$Z$\}-min\{$Z$\}. \\ \midrule
2 & LT & Outer & \begin{tabular}[c]{@{}l@{}}Max of $X$ and $Y$ extents in cluster, Beam 7. This is max\{extent\{$X$\},extent\{$Y$\}\}.\\ Extent \{$X$\} = max\{$X$\}-min\{$X$\}. Extent \{$Y$\} = max\{$Y$\}-min\{$Y$\}.\end{tabular} \\ \midrule
3 & LT & Outer & \begin{tabular}[c]{@{}l@{}}Max of $X$ and $Y$ extents in cluster, Beam 5. This is max\{extent\{$X$\},extent\{$Y$\}\}.\\ Extent \{$X$\} = max\{$X$\}-min\{$X$\}. Extent \{$Y$\} = max\{$Y$\}-min\{$Y$\}.\end{tabular} \\ \midrule
4 & HT & Outer & Max\{$Z$ in cluster - LiDAR height\}. $Z$ is vertical (height) of LiDAR return. \\ \midrule
5 & LT & Outer & Extent of $Z$ in cluster. Extent\{$Z$\}= max\{$Z$\} - min\{$Z$\}. \\ \midrule
6 & LT & Inner & Number of valid points in cluster, Beam 7. \\ \midrule

7 & LT & Inner & \begin{tabular}[c]{@{}l@{}}Max of $X$ and $Y$ extents in cluster, Beam 6. This is max\{extent\{$X$\},extent\{$Y$\}\}.\\ Extent \{$X$\} = max\{$X$\}-min\{$X$\}. Extent \{$Y$\} = max\{$Y$\}-min\{$Y$\}.\end{tabular} \\ \midrule
8 & LT & Outer & Number of points in cluster, Beam 5. \\ \midrule
9 & LT & Inner & Extent of $X$ in cluster. Extent\{$X$\}= max\{$X$\} - min\{$X$\}. \\ \midrule
10 & LT & Inner & Number of points in cluster, Beam 4. \\ \midrule
11 & LT & Inner & Number of points in cluster, Beam 5. \\ \midrule
12 & LT & Outer & Number of points in cluster, Beam 6. \\ \midrule
13 & HT & Inner & Number of points in cluster, Beam 6. \\ \midrule
14 & LT & Inner & \begin{tabular}[c]{@{}l@{}}Max of $X$ and $Y$ extents, Beam 5. This is max\{extent\{$X$\},extent\{$Y$\}\}.\\ Extent \{$X$\} = max\{$X$\}-min\{$X$\}. Extent \{$Y$\} = max\{$Y$\}-min\{$Y$\}.\end{tabular} \\ \midrule
15 & LT & Outer & Number of points in cluster divided by the cluster radius in Beam 5. \\ \midrule
16 & LT & Inner & Extent of $X$ in cluster. Extent \{$X$\} = max\{$X$\}-min\{$X$\}. \\ \midrule

17 & HT & Inner & Number of points in cluster, Beam 7. \\ \midrule
18 & LT & Inner & Number of points in cluster. \\ \midrule
19 & LT & Outer & Number of points in cluster. \\ \midrule
20 & LT & Outer & Extent of $Z$ in cluster. Extent \{$Y$\} = max\{$Y$\}-min\{$Y$\}. \\ \bottomrule
\end{tabular}
\end{table}\unskip

\subsubsection{SVM LiDAR Beacon Detection}
\label{subsubsubsec:svmlidar}

\begin{figure}[H]
\centering
\caption{SVM 2D example. The margin is the distance from the $D = -1$ to $D = 1$ hyperplanes.}
\label{fig:SVM}
\includegraphics[width=4in]{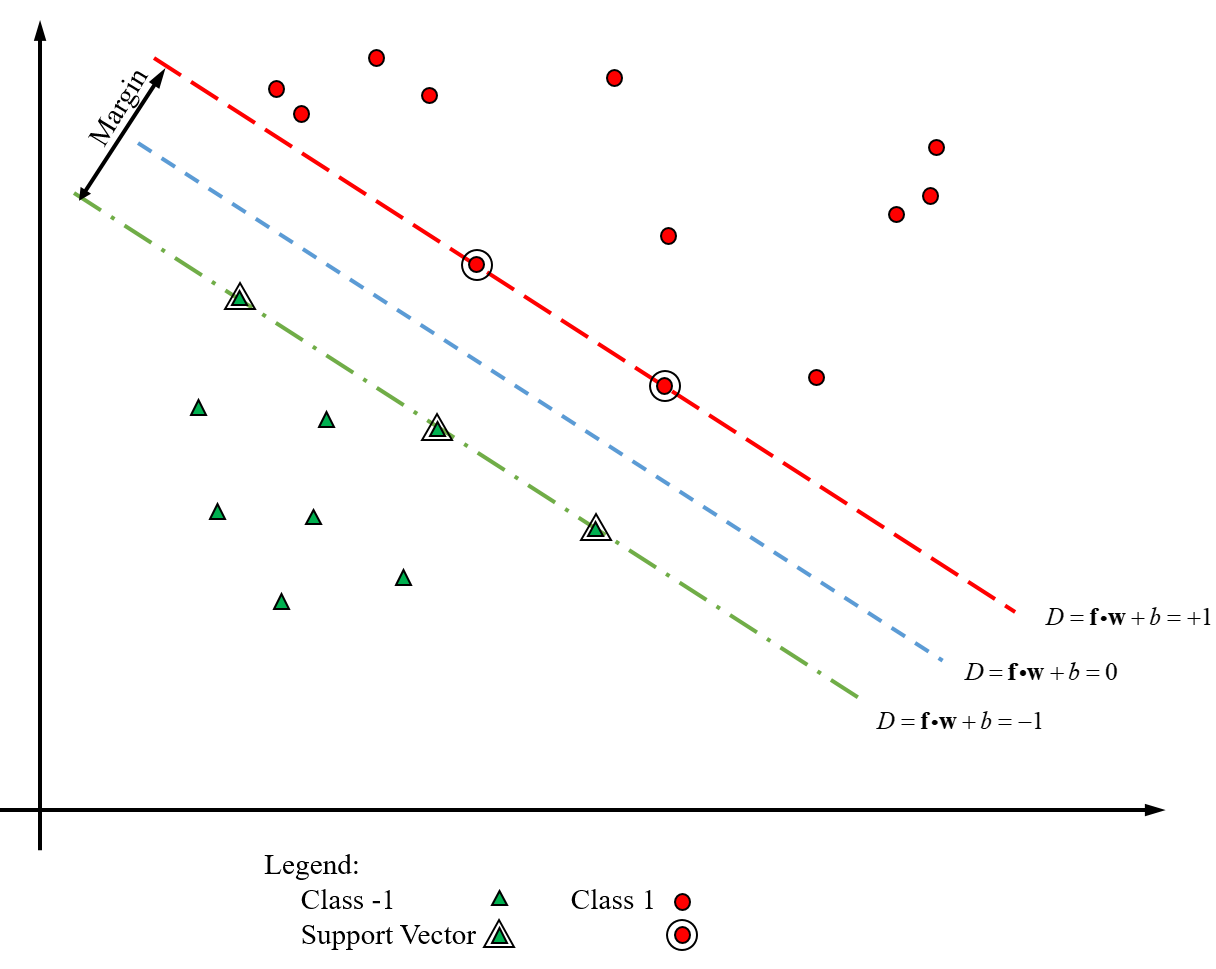}
\end{figure}

To detect the beacons in the 3D LiDAR point cloud, we use a~linear SVM \citep{Vapnik1998}, which makes a~decision based on a~learned linear discriminant rule operating on the hand-crafted features we have developed for differentiating beacons. Herein, liblinear was used to train the SVM \citep{Fan2008}. A linear SVM operates by finding the optimal linear combination of features that best separates the training data. If~the $M$ testing features for a~testing instance are given by ${\bf{f}} = {\left[ {{f_1},{f_2}, \cdots ,{f_M}} \right]^T}$, where the superscript $T$ is a~vector transpose operator, then the SVM evaluates the discriminant: 
\begin{equation}
 D = {\bf{f}} \cdot {\bf{w}} + b,
\label{eqn:SVMDiscriminant}
\end{equation}
where $\textbf{w}$ is the SVM weight vector and $b$ is the SVM bias term. Herein, $M$ was chosen to be 20. Equation (\ref{eqn:SVMDiscriminant}) applies a~linear weight to each feature and adds a~bias term. The discriminant is optimized during SVM training. The SVM optimizes the individual weights in the weight vector to maximize the margin and provide the best overall discriminant capability of the SVM. The bias term lets the optimal separating hyperplane drift from the origin. {Figure} \ref{fig:SVM} shows an example case with two features. \mbox{In this case}, the $D = 0$ hyperplane is the decision boundary. The $D = +1$ and $D = -1$ hyperplanes are determined by the support vectors. The SVM only uses the support vectors to define the optimal~boundary.
%
%
\unskip

The SVM was trained with beacon and non-beacon data extracted over multiple data collections. There were 13,190 beacon training instances and 15,209 non-beacon training instances. The test data had 12,084 beacons and 5666 non-beacon instances.

In this work, $M = 20$ features were chosen. Each feature is first normalized by subtracting the mean of the feature values from the training data and dividing each feature by four times the standard deviation plus $10^{-5}$ (in order to prevent division by very small numbers). Subtracting the mean value centers the feature probability distribution function (PDF) around zero. Dividing by four times, the standard deviation maps almost all feature values into the range $\left[ -1,1 \right]$, because most of the values lie within $\pm 4{\sigma _k}$. The normalized features are calculated using:
\begin{equation}
 {f'_k} = \frac{{{f_k} - {\mu _k}}}{{4{\sigma _k} + {{10}^{ - 5}}}}
\label{eqn:FeatureNormalization}
\end{equation}
where ${\mu _k}$ is the mean value of feature $k$ and ${\sigma _k}$ is the standard deviation of feature $k$ for $k = 1,2, \cdots, M$. The mean and standard deviation are estimated from the training data and then applied to the test~data. The final discriminant is computed using Equation (\ref{eqn:SVMDiscriminant}) where the feature vector is the normalized feature~vector.

\begin{figure}[H]
\centering
\caption{Optimal feature weights chosen by the SVM training optimizer.}
\label{fig:OptimalSVMWeights}
\includegraphics[width=5in]{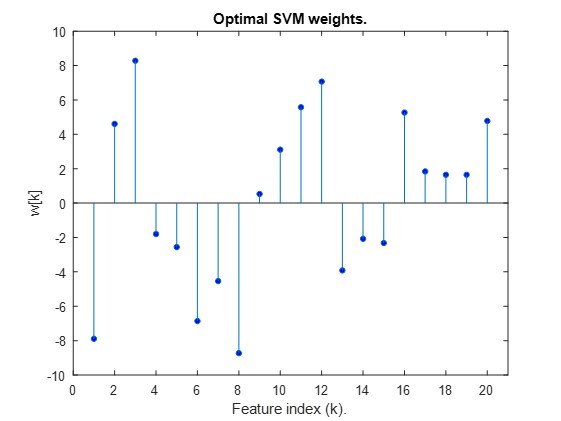}
\end{figure}

The optimal feature weighs are shown in Figure \ref{fig:OptimalSVMWeights}. Since all the features are normalized, Features 1, 3 and 8 have the most influence on the final discriminant function.

If the discriminant $D \le 0$, then the object is declared a~beacon. Otherwise, it is declared a~non-beacon (and ignored by the LiDAR beacon detection system). Once the SVM is trained, implementing the discriminant given in Equation (\ref{eqn:SVMDiscriminant}) is trivial and uses minimal processing time. 

\begin{figure}[H]
\centering
\caption{LiDAR discriminant values.}
\label{fig:LiDARDiscrimiantValues}
\includegraphics[width=4in]{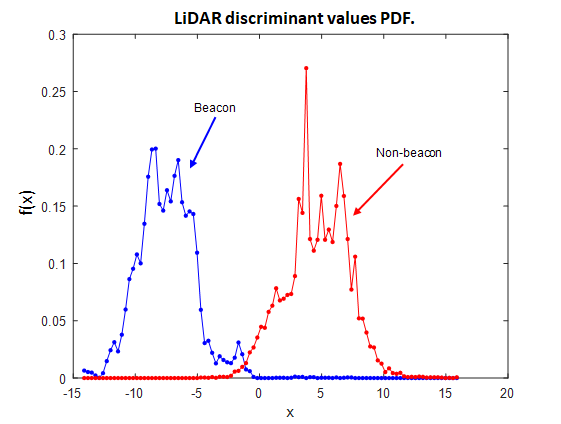}
\end{figure}

Figure \ref{fig:LiDARDiscrimiantValues} shows the {PDF} of the LiDAR discriminant values for beacons and non-beacons, respectively. The PDFs are nearly linearly separable. The more negative the discriminant values, the more the object is beacon-like.

\subsection{Mapping}
This section discusses the camera detection mapping from bounding box coordinates to distance and angle. 
It also discusses the mapping of LiDAR discriminate value from the range $[-\infty,\infty]$ to $[0, 1]$.

\subsubsection{Camera Detection Mapping}
\label{subsec:LiDARCameraDecisionFusion}
\begin{figure}[H]
\centering
\caption{Detection mapping for the camera.}
\label{fig:nn}
\includegraphics[width=4in]{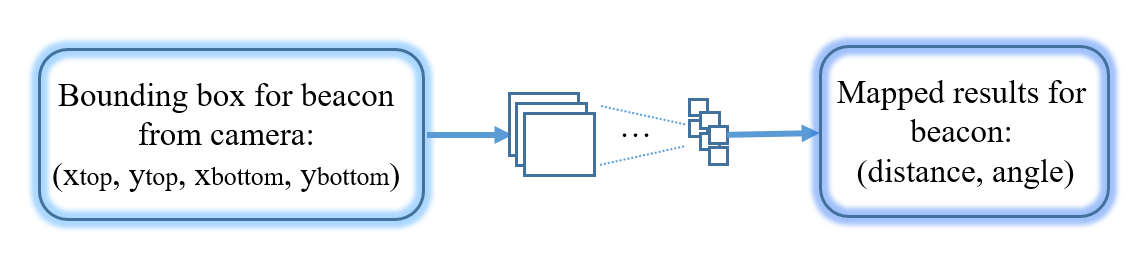}
\end{figure}

The LiDAR and the camera both provide information on object detection, although in entirely different coordinate spaces. The LiDAR provides accurate angle and range estimates with the discriminant value offering a~rough metric of confidence. The camera processing algorithms return bounding box coordinates, class labels and confidence values in the range $[0,1]$. However, effective fusion requires a~common coordinate space. Therefore, a~mapping function is required to merge these sensor measurements into a~shared coordinate space.
To build this mapping function, the camera and LiDAR were mounted to the industrial vehicle, and the LiDAR's accurate position measurements, alongside the camera's bounding boxes, were used to collect training data. {These data were} then used to train a~neural network. The end result was a~neural network that could project the camera's bounding boxes into a~range and angle estimate in the LiDAR's coordinate frame, as seen in Figure \ref{fig:nn}. In this figure, the middle portion simply represents the bounding boxes from the camera system, and the smaller boxes before the mapped results represent the NN mapper.

\begin{figure}[H]
\centering
\caption{NN structure for detection mapping for the camera.}
\label{fig:nndetail}
\includegraphics[width=6in]{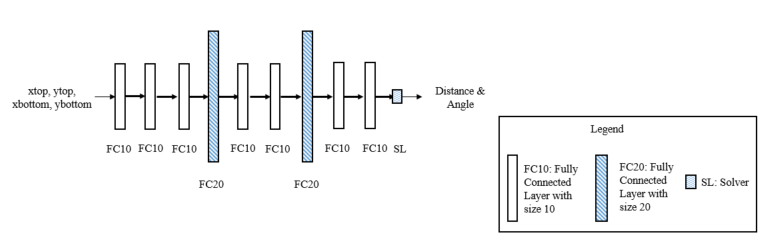}
\end{figure}

The detailed NN structure is shown in Figure \ref{fig:nndetail}. It consists of ten fully-connected layers: eight layers of ten neurons and two layers of twenty neurons. We choose this structure because of its simplicity to compute while outperforming other linear and non-linear regression methods. 
  
\subsubsection{LiDAR Discriminate Value Mapping}
With the camera bounding boxes mapped to LiDAR range angle, there still remains the mapping of the discriminate value from LiDAR and confidence scores from the camera into a~shared space. For this, the $[0,1]$ range of the camera confidence offers a~more intuitive measure than the LiDAR's $[-\infty,\infty]$ range for the discriminate value. Thus, the mapping function for this component will map the LiDAR discriminate to a~$[0,1]$ space for fusion with the camera. To do this, a~logistic sigmoid of the form
%
%
\begin{equation}
\label{eqn:Sigmoid}
 \begin{split}
   f(x) = \frac{1}{1+e^{-\alpha*x}}
 \end{split}
\end{equation}
{was used}. The discriminate value is multiplied by a~gain term, $\alpha$. 
The end result is a~function that maps the discriminate value to a~pseudo-confidence score with a~range similar enough to the camera's confidence score for effective fusion.

\subsection{Detection Fusion} \label{subsec:detectionfusion}

In this section, we discuss the fusion algorithm and hyperparameter optimization.
\subsubsection{Fusion Algorithm}
With these two mapping components, a~full pipeline can now be established to combine the LiDAR and camera data into forms similar enough for effective fusion. Using distance and angle information from both the camera and LiDAR, we can correlate LiDAR and camera detections together. Their confidence scores can then be fused as shown in Algorithm \ref{alg:fusion}.
  
\vspace{8pt}

\begin{algorithm}[H]
\textbf{\\}
\KwIn{Detection from LiDAR in the form of [distance, angle, pseudo-confidence score].}
\KwIn{Detection from camera in the form of [distance, angle, confidence score].}
\KwIn{Angular threshold: $A$ for angle difference between fused camera and LiDAR detection.}
\KwIn{Confidence threshold: $C$ for final detection confidence score threshold.}

\textbf{\\}
\KwOut{Fused detection in the form of [distance, angle, detection confidence].}

\textbf{\\}
\For{each detection from camera}{
  \eIf{a corresponding LiDAR detection, with angle difference less than $A$, exists}
  {
   Fuse by using the distance and angle from LiDAR, and combine confidence scores using fuzzy logic to determine the fused detection confidence.
  }
  {
   Create a~new detection from camera, using the distance, angle and confidence score from camera as the final detection result.
  }
}

\textbf{\\}
\For{each LiDAR detection that does not have a~corresponding camera detection}{ 
  Create a~new detection from LiDAR, using the distance, angle and confidence score from LiDAR as the final detection result.
}

\textbf{\\}
\For{each fused detection}{
  Remove detections with final confidences below the $C$ confidence threshold.
}

\textbf{\\}
\text{Return fused detection results.}
\caption{{Fusion of LiDAR and Camera detection.} 
\label{alg:fusion}}

\end{algorithm}

\vspace{8pt}

\begin{figure}[H]
\centering
\begin{subfigure}[b]{0.3\textwidth}
\includegraphics[width=1.0\textwidth]{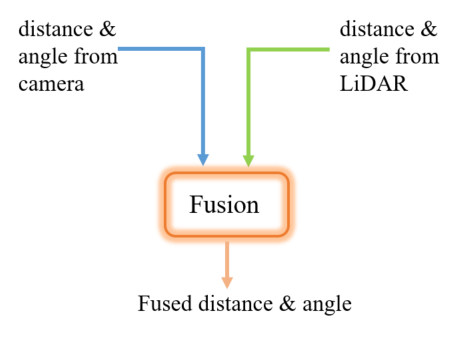}
\caption{}
\label{subfig:fusionda}
\end{subfigure}
\begin{subfigure}[b]{0.3\textwidth}
\includegraphics[width=1.0\textwidth]{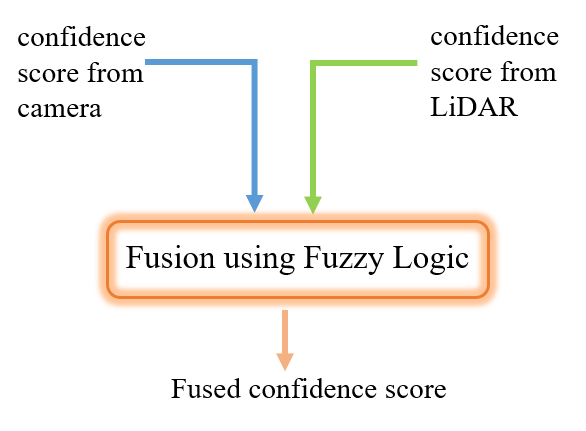}
\caption{}
\label{subfig:fusioncs}
\end{subfigure}
\caption{(\textbf{a}) Fusion of distances and angles. (\textbf{b}) Fusion of confidence scores. Fusion of distances, angles and confidence scores from the camera and LiDAR.}
\label{fig:twofusion}
\end{figure}

In this algorithm, the strategy for fusion of the distance and angle estimates is different from the fusion of confidence scores, which is shown in Figure \ref{fig:twofusion}.
In the process of distance and angle fusion, for each camera detection, when there is a~corresponding detection from LiDAR, we will use the distance and angle information from LiDAR, as LiDAR can provide much more accurate position estimation. When there is no corresponding LiDAR detection, we will use the distance, angle and confidence score from the camera to create a~new detection. For each detection from LiDAR that does not have corresponding camera detection, we will use the distance, angle and confidence score from LiDAR as the final result.
For confidence score fusion, we use fuzzy logic to produce a~final confidence score when there is corresponding camera and LiDAR detection. The fuzzy rules used in the fuzzy logic system are as follows:
\begin{enumerate}
  \item If the LiDAR confidence score is high, then the probability for detection is high.
  \item If the camera confidence score is high, then the probability for detection is high.
  \item If the LiDAR confidence score is medium, then the probability for detection is medium.
  \item If the LiDAR confidence score is low and the camera confidence score is medium, then the probability for detection is medium.
  \item If the LiDAR confidence score is low and the camera confidence score is low, then the probability for detection is low.
\end{enumerate}

The fuzzy rules are chosen in this way because of the following reasons: 
First, if either the LiDAR or the camera shows a~high confidence in their detection, then the detections are very likely to be true. This is the reason for Rules 1 and 2: the LiDAR is robust at detection up to about 20 m. Within that range, the confidence score from the LiDAR is usually high or medium. Beyond that range, the value will drop to low. This is the reason for Rule 3. Third, beyond the 20-m detection range, the system will solely rely on the camera for detection. This is the reason for Rules 4 and 5.

\begin{figure}[H]
\centering
\begin{subfigure}[b]{0.43\textwidth}
\includegraphics[width=\textwidth]{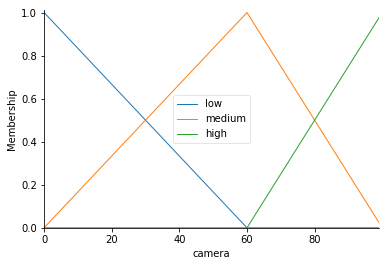}
\caption{}
\label{subfig:fcam}
\end{subfigure}
\begin{subfigure}[b]{0.45\textwidth}
\includegraphics[width=\textwidth]{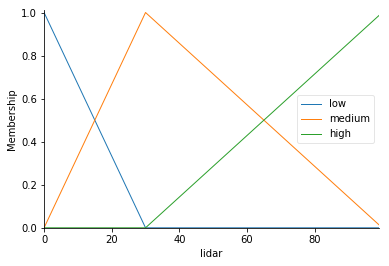}
\caption{}
\label{subfig:flid}
\end{subfigure}
\begin{subfigure}[b]{0.45\textwidth}
\includegraphics[width=\textwidth]{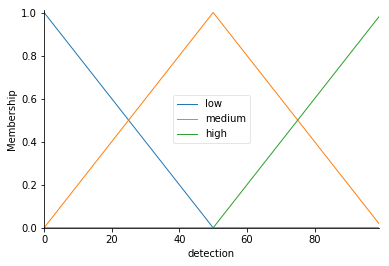}
\caption{}
\label{subfig:fdet}
\end{subfigure}
\begin{subfigure}[b]{0.45\textwidth}
\includegraphics[width=\textwidth]{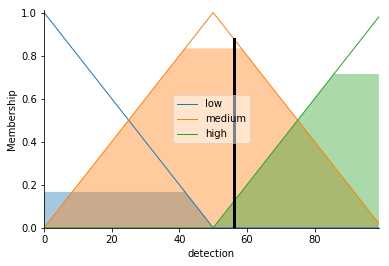}
\caption{}
\label{subfig:fres}
\end{subfigure}
\caption{(\textbf{a}) Fuzzy membership function for camera. (\textbf{b}) Fuzzy membership function for LiDAR. (\textbf{c})~Fuzzy membership function for detection. (\textbf{d}) Fuzzy logic output. Combination of confidence scores from the camera and LiDAR using fuzzy logic. Figure best viewed in color.}
\label{fig:fuzzylogic}
\end{figure}

Figure \ref{fig:fuzzylogic} shows the process of using fuzzy logic to combine the confidence scores from the camera and LiDAR. The camera input, LiDAR input and detection output are fuzzified into fuzzy membership functions as shown in Figures \ref{fig:fuzzylogic}a to \ref{fig:fuzzylogic}c. The output is shown in Figure \ref{fig:fuzzylogic}d. In this example, when~the confidence score from the LiDAR is 80 and the score from the camera is 20, by applying the five pre-defined fuzzy rules, the output detection score is around 56.

\unskip

%
%

\subsubsection{Hyperparameter Optimization}
\begin{figure}[H]
\centering
\caption{Fusion of confidence scores with two hyperparameters.}
\label{fig:para}
\includegraphics[width=4in]{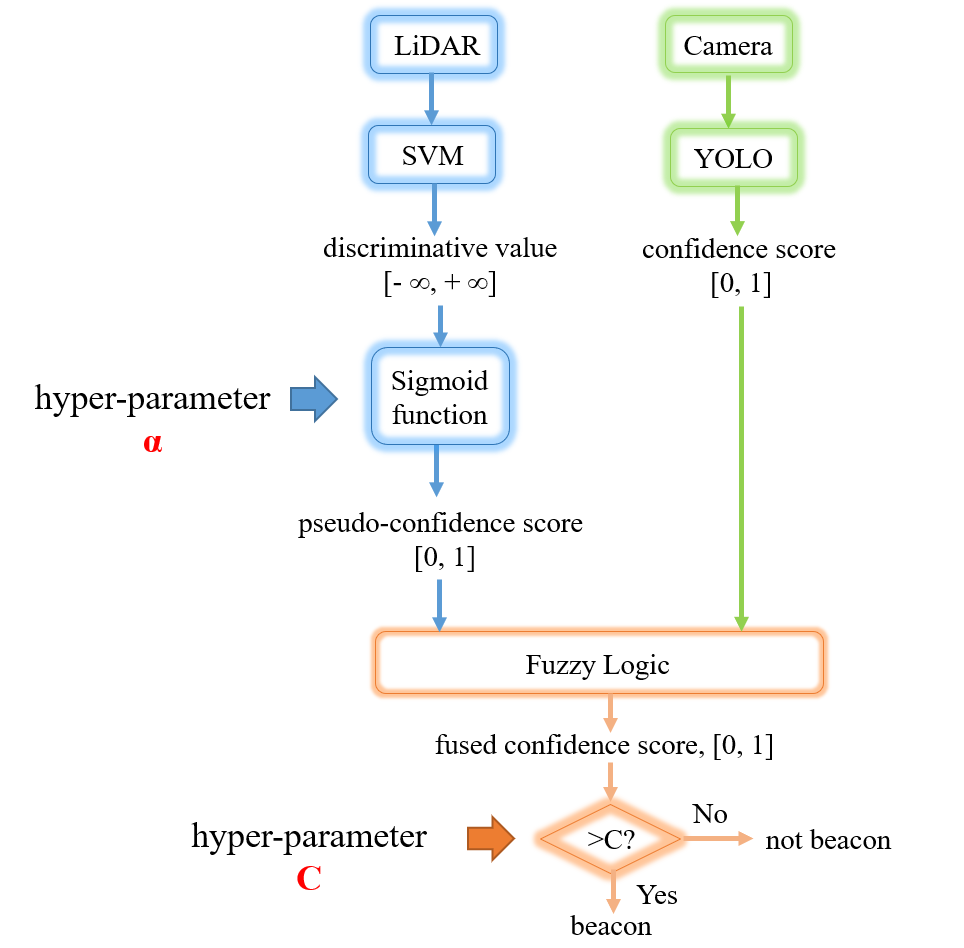}
\end{figure}

In machine learning, the value of hyperparameter is set before the learning process. 
In~Equation~(\ref{eqn:Sigmoid}), $\alpha$ controls the LiDAR sigmoid-like squashing function and is one of the hyperparameters to be decided.
In Algorithm \ref{alg:fusion}, there is another hyperparameter threshold $C$ for the final detection. Figure \ref{fig:para} shows where these two hyperparameters affect the fusion process. For the optimization of hyperparameters, some commonly-used approaches include marginal likelihood \citep{zorzi2017sparse}, grid search \citep{chicco2017ten, hsu2003practical}, evolutionary optimization \citep{olson2016automating}, and many others. Herein, we utilized grid search. This method is widely used in machine learning system as shown in \citep{chicco2017ten, hsu2003practical}. In grid search, the two hyperparameters are investigated by selecting candidate values for each, then generating a~2D matrix of their cross~product.

As we have two hyperparameters $\alpha$ and $C$ to be tuned, we use the Kolmogorov--Smirnov test \citep{chakravarty1967handbook}, also known as the KS-test, to determine which hyperparameter values to choose. In our application, the KS-test is to find the $\alpha$ and $C$ that maximize the value of TPR minus FPR. As we want to maximize TPR and minimize FPR, the purpose of the KS-test is to find the balance between the two metrics.

\begin{figure}[H]
\centering
\begin{subfigure}[b]{0.7\textwidth}
\includegraphics[width=\textwidth]{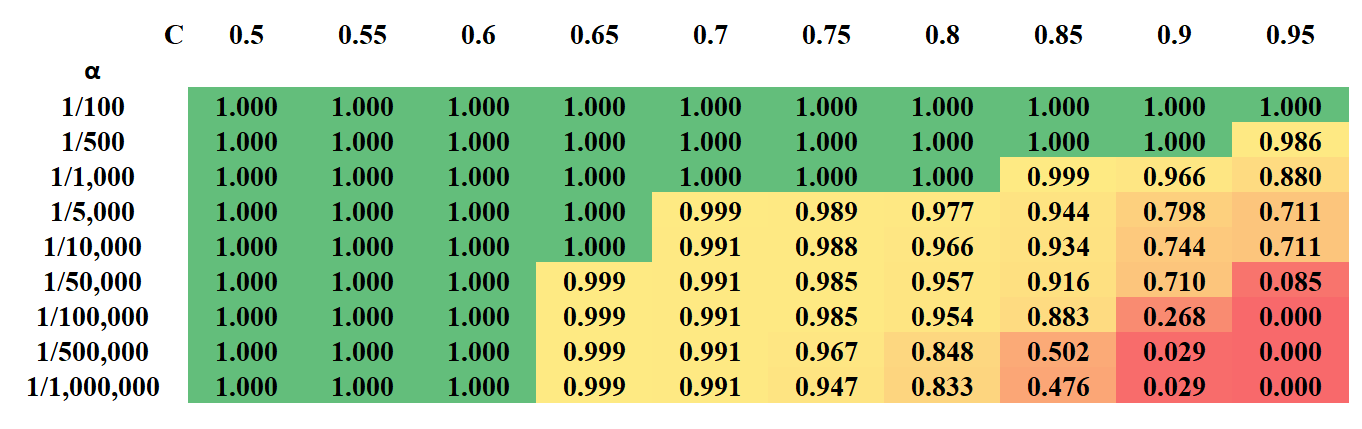}
\caption{}
\label{subfig:tpr}
\end{subfigure}

\begin{subfigure}[b]{0.7\textwidth}
\includegraphics[width=\textwidth]{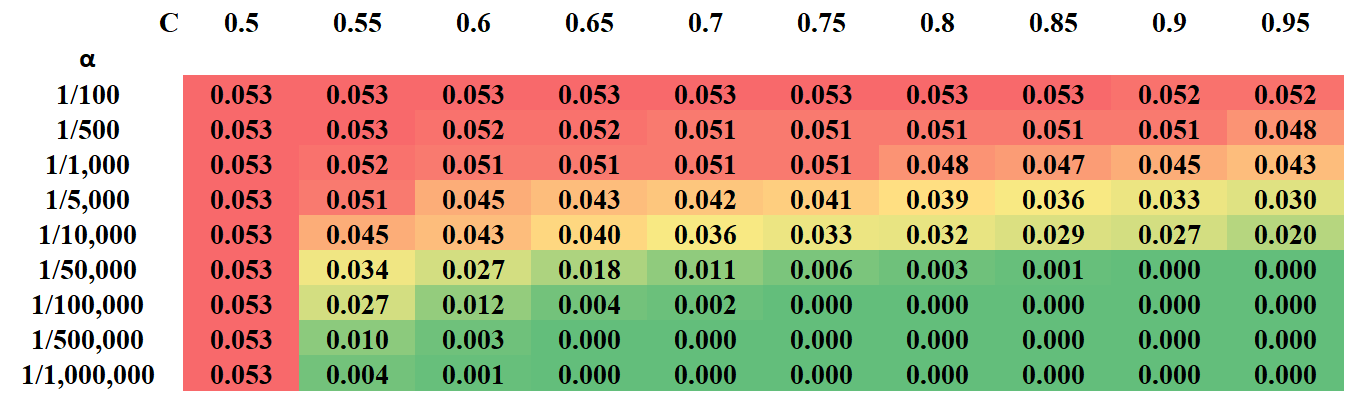}
\caption{}
\label{subfig:fpr}
\end{subfigure}

\begin{subfigure}[b]{0.7\textwidth}
\includegraphics[width=\textwidth]{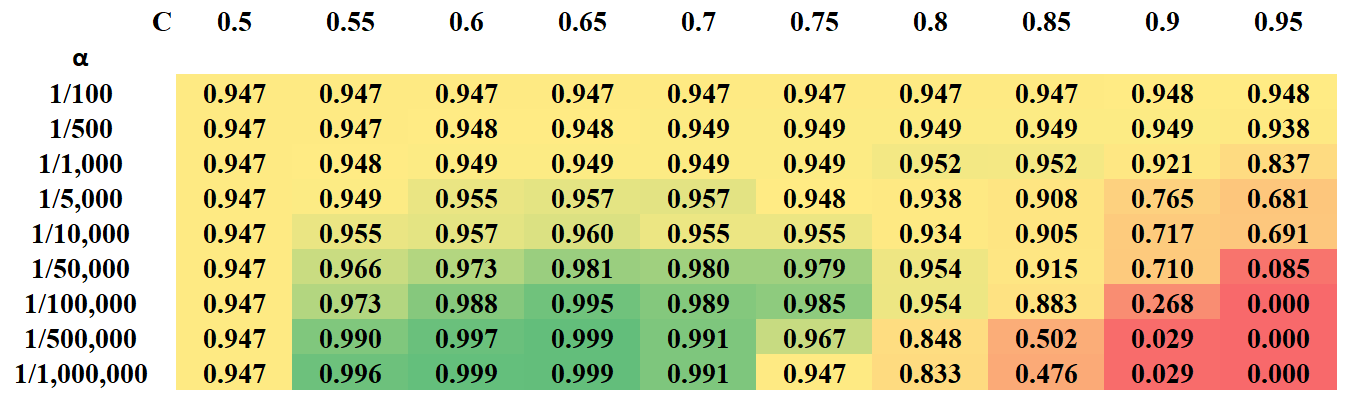}
\caption{}
\label{subfig:ks}
\end{subfigure}

\caption{(\textbf{a}) True positive rate (TPR) changing with $\alpha$ and $C$. (\textbf{b}) False positive rate (FPR) changing with $\alpha$ and $C$. (\textbf{c}) KS-test results changing with $\alpha$ and $C$. Heat maps demonstrating the TPR, FPR and KS-test results changing with $\alpha$ and $C$. Best viewed in color.}
\label{fig:heatmap}
\end{figure}

The values chosen for $\alpha$ are $\frac{1}{100}$, $\frac{1}{500}$, $\frac{1}{1000}$, $\frac{1}{5000}$, $\frac{1}{10,000}$, $\frac{1}{50,000}$, $\frac{1}{100,000}$, $\frac{1}{500,000}$ and $\frac{1}{1,000,000}$, while the values for $C$ are $0.5$, $0.55$, $0.6$, $0.65$, $0.7$, $0.75$, $0.8$, $0.85$, $0.9$ and $0.95$. We use 3641 testing LiDAR and camera pairs, and the resulting heat maps for TPR, FPR and the KS-test results are shown in Figure \ref{fig:heatmap}. 
As~we want the TPR to be high, in Figure \ref{fig:heatmap}a, the green area shows where the highest TPR is, while the red are is where the lowest TPR is.
Furthermore, we want the FPR to be low, in Figure \ref{fig:heatmap}b; the lower values are in the green area, while the higher values are in the red area.
To balance between TPR and FPR, Figure~\ref{fig:heatmap}c shows the KS-test results, in other words the difference between TPR and FPR. We want to choose the $\alpha$ and $C$ related to higher values (green area) in this heat map. Based on the KS-test, {there are three pairs of} $\alpha$ and $C$ {values that can be chosen}, and we choose $\alpha$ to be $\frac{1}{500,000}$ and $C$ to be 0.65 in our experiment. 

%
%

\section{Experiment}
\label{sec:experiment}

\begin{figure}[H]
\centering
\begin{subfigure}[b]{0.25\textwidth}
\includegraphics[width=0.7\textwidth]{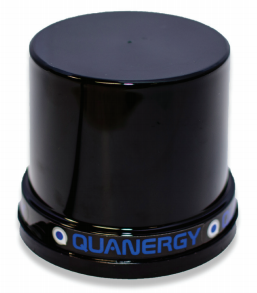}
\caption{}
\label{subfig:Quanergy_lidar}
\end{subfigure}
\begin{subfigure}[b]{0.25\textwidth}
\includegraphics[width=\textwidth]{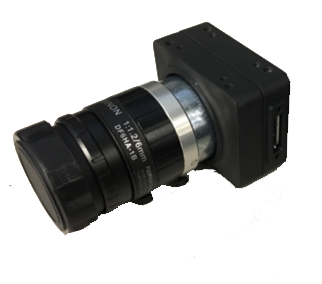}
\caption{}
\label{subfig:cam}
\end{subfigure}
\caption{(\textbf{a}) Quanergy M8-1 LiDAR. (\textbf{b}) FLIR Chameleon3 USB camera with a Fujinon 6-mm lens. The sensors used in our experiments.}
\label{fig:equip}
\end{figure}

In our experiments, the LiDAR we use is a~Quanergy M8-1 LiDAR, shown in Figure \ref{fig:equip}a. It has eight beams with a~vertical spacing of approximately $3^{\circ}$. The camera images are collected by an~FLIR Chameleon3 USB camera with a Fujinon 6-mm lens, as shown in Figure \ref{fig:equip}b. 

%
%

%
%

\subsection{Data Collections Summary}
To collect data and test our system in a~relevant environment, we needed an open outdoor space without much clutter and interference from random objects. We chose an industrial setting that is very similar to the scenario where we will implement our system. The location has {an} office building and a~large, relatively flat concrete area adjacent to the building. In the environment, there are a~few parked cars, a~trailer, a~couple of gas storage tanks, etc.

Twenty extensive data collections were conducted from August 2017 to March 2018. These tests are designed to collect training data for our system and testing data to check the system's performance and to tune the system parameters. Many tests consisted of beacons placed in a~random position in the sensors' FOV. Other tests had beacons in known locations in order to assess the LiDAR and camera accuracies for estimating the range and distance. Later tests used the LiDAR as the ``ground truth'', since it is highly accurate at estimating range and distance.

Most of the initial tests had the industrial vehicle stationary. Later, we performed tests with the industrial vehicle moving. For the initial tests where we needed to know beacon locations, we~marked the different ground distances and angles. The ground point straight below the LiDAR is used as the 0-m distance, and the $0^{\circ}$ angle is directly in front of the industrial vehicle. From there, we marked the distance from 3 m at each meter interval up to 40 m across the $0^{\circ}$ angle line. The $0^{\circ}$ angle is directly in front of the industrial vehicle. We also marked the $-20^{\circ}, -15^{\circ}, -10^{\circ}, -5^{\circ}, 0^{\circ}, 5^{\circ}, 10^{\circ}, 15^{\circ}$ and $20^{\circ}$ lines for beacon location. In this way, we constructed a~dataset of beacons with labeled ground~truths. For~static beacon testing, we placed the beacon in both known and unknown locations and recorded the~data. 

For the stationary vehicle testing, we set up several tests to evaluate different parts of the system. For instance, a~beacon was placed on a~very short wooden dolly attached to a~long cord. This allowed the beacon to be moved while the people moving it were not in the sensors' field of view. This allowed the beacon to be placed at a~known distance along the $0^{\circ}$ angle line and drug toward the vehicle to collect data at continuous locations. We also did similar tests along the $-40^{\circ}, -20^{\circ}, -10^{\circ}, 10^{\circ}, 20^{\circ}$ and $40^{\circ}$ angle lines. Finally, we collected data where the beacon was drug from the $-40^{\circ}$ angle to all the way across the $40^{\circ}$ angle at roughly fixed distances. These data collections gave us a~large amount of data to use in training and testing our system.

We also performed data collections with humans. The retro-reflective stripes on the vests present bright returns to the LiDAR, which may cause false positives. To measure these effects, we had different people wear a~highly reflective vest and stand in front of the LiDAR and cameras in a~known location. Data were gathered with the person in multiple orientations facing toward and away from the camera. This test was also performed without the reflective safety vests. Other tests had people walking around in random directions inside the sensors' FOV. The data collected here {were} part of the non-beacon data used to train the SVM. 

In order to examine interference when people and beacons were also in the scene, we had people wearing a~vest standing close to the beacon in known locations. Then, we recorded the data, which helped the system to distinguish the beacon and people even if in close contact. Furthermore, we also wanted some dynamic cases of beacons and people in close proximity. To gather these data, the beacon was pulled around using the rope and dolly, while a~second person moved around the beacon. 

There was also a~series of data collections with the vehicle driving straight at or just past a~beacon or another object, to test that the sensor system was stable under operating conditions. The vehicle was driven at a~constant speed, accelerating and braking (to make the vehicle rock) and in a~serpentine manner. Finally, we performed many data collections with the vehicle driving around with different obstacles and beacons present. This was mainly to support system integration and tune the MPC controller and actuators. All of these collections provided a~rich dataset for system analysis, tuning and performance evaluation.

\subsection{Camera Detection Training}
Training images for YOLO also include images taken by a~Nikon D7000 camera and images from PASCAL VOC dataset \citep{everingham2010pascal}. The total number of training images is around 25,000. 

The network structure of YOLO is basically the same as the default setting. One difference is the number of filters in the last layer. This number is related to the number of object categories we are trying to detect and the number of base bounding boxes for each of the grid sections into which YOLO divides the image. For example, {we can divide} each image into 13 by 13 grids. For each grid, YOLO predicts five bounding boxes, and for each bounding box, there are $5+N$ parameters ($N$ represents the number of categories to be predicted). As a~result, the last layer filter size is $13 \times 13 \times 5 \times (5 + N)$.
Another difference is that we change the learning rate to $10^{-5}$ to avoid divergence. 

\subsection{LiDAR Detection Training}

The LiDAR data {were} processed through a~series of steps. We implemented a~ground clutter removal algorithm to keep ground points from giving us false alarms, to avoid high reflections from reflective paints on the ground and to reduce the size of our point cloud. Clusters of points are grouped into distinct objects and are classified using an~SVM. The SVM is trained using a~large set of data that have beacons and non-beacon objects. The LiDAR data {were} divided into two disjoint sets: training and~testing. In the training dataset, there were 13,190 beacons and 15,209 non-beacons. The~testing dataset contained 5666 beacons and 12,084 non-beacons. The confusion matrices for training and testing are shown in Tables \ref{tbl:LiDARTrainConfusionMatrix} and \ref{tbl:LiDARTestConfusionMatrix}. The results show good performance for the training and testing datasets. Both datasets show over $99.7\%$ true positives (beacons) and around $93\%$ true negatives (non-beacons). 

%
%
\begin{table}[H]
\centering
\caption{LiDAR training confusion matrix.}
\label{tbl:LiDARTrainConfusionMatrix}
\begin{tabular}{lcc}
\toprule
 & \multicolumn{1}{l}{\textbf{Beacon}} & \multicolumn{1}{l}{\textbf{Non-Beacon}} \\ \midrule
\multicolumn{1}{l}{{Beacon}} & 13,158 & 32 \\ \midrule 
\multicolumn{1}{l}{{Non-Beacon}} & 610 & 14,599 \\ \bottomrule
\end{tabular}
\end{table}\unskip

%
%
\begin{table}[H]
\centering
\caption{LiDAR testing confusion matrix.}
\label{tbl:LiDARTestConfusionMatrix}
\begin{tabular}{lcc}
\toprule
 & \multicolumn{1}{l}{\textbf{Beacon}} & \multicolumn{1}{l}{\textbf{Non-Beacon}} \\ \midrule
\multicolumn{1}{l}{{Beacon}} & 5653 & 13 \\ \midrule 
\multicolumn{1}{l}{{Non-Beacon}} & 302 & 11,782 \\ \bottomrule
\end{tabular}
\end{table}

%
%
\section{Results and Discussion} \vspace{-6pt}
\label{sec:results}

\subsection{Camera Detection Mapping Results}
Camera and LiDAR detections generate both different types of data and data on different~scales. The LiDAR generates range and angle estimates, while the camera generates bounding boxes. The~LiDAR discriminant score can be any real number, while the camera confidence score is a~real number in the range $[0,1]$. The data from camera detection are first mapped into the form of distance and range, the same form as the data reported from LiDAR. We can accomplish this conversion fairly accurately because we know the true size of the beacon. 
The data collected for mapping cover the entire horizontal FOV of the camera, which is $-20^{\circ}$ to $20^{\circ}$ relative to the front of the industrial vehicle. The~distance covered is from 3 to 40 m, which is beyond the maximum detection range of our LiDAR detection algorithm. We collected around 3000 pairs of camera and LiDAR detection for training of the mapping NN and 400 pairs for testing. 

\begin{figure}[H]
\centering
\caption{One sample frame with three different bounding boxes.}
\label{fig:regress}
\includegraphics[width=3in]{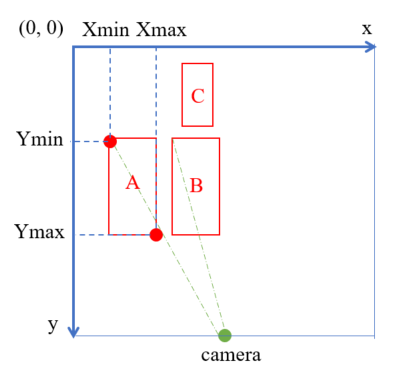}
\end{figure}

\begin{figure}[H]
\centering
\begin{subfigure}[b]{0.5\textwidth}
\includegraphics[width=\textwidth]{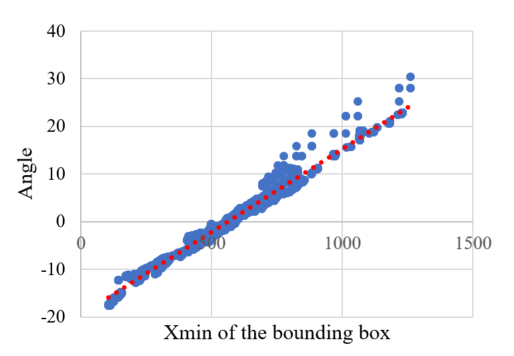}
\caption{Bounding box $Xmin$ from the camera vs. the angle from LiDAR. Xmin 
units are pixels, and angle units are~degrees.}
\label{subfig:a}
\end{subfigure}
\begin{subfigure}[b]{0.5\textwidth}
\includegraphics[width=\textwidth]{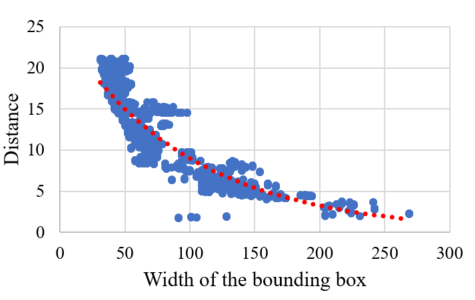}
\caption{Bounding box $Xmin$ from the camera vs. the angle from LiDAR. The bounding box width is pixels, and the distance is~meters.}
\label{subfig:d}
\end{subfigure}
\caption{Scatter plots of camera/LiDAR training data for fusion. (\textbf{a}) Camera bounding box $Xmin$ vs. LiDAR estimated angle. (\textbf{b}) Camera bounding box width vs. LiDAR estimated distance. }
\label{fig:trending}
\end{figure}

From Figure~\ref{fig:regress}, we observe that the larger the $x$-coordinate, the larger the angle from the camera, and the larger the size of the bounding box, the nearer the object to the camera. From our training~data, we see that the bounding box's $x$-coordinate has an approximately linear relationship with the LiDAR detection angle, as shown in Figure~\ref{fig:trending}a. We also observe that the size of the bounding box from the camera detection has a~nearly negative exponential relationship with the distance from LiDAR detection as shown in Figure~\ref{fig:trending}b. To test these relationships, we ran least squares fitting algorithms to estimate coefficients for a~linear fit between the [$Xmin$, $Xmax$] of the camera bounding box and the LiDAR angle and for an exponential fit between the [$Width$, $Height$] and the LiDAR estimated~distance.

Herein, we compare the NN and linear regression result methods for angle estimation, and we also compare the results using NN and exponential curve fitting method for distance. The results are shown in Table~\ref{tbl:MappingAnalysis}. The table shows the mean squared error (MSE) using both the NN and the linear regression for the angle estimation, as well as the MSE for distance estimation. The table also reports the coefficient of determination, or $r^{2}$ scores \citep{draper2014applied}. The $r^{2}$ scores show a~high measure of goodness of fit for the NN approximations. We find that the NN provides more accurate predictions on both distance and angle. 

\begin{table}[H]
\centering
\caption{Mapping analysis. The best results are shown in {\textbf{bold}}.}
\label{tbl:MappingAnalysis}
\begin{tabular}{ccccc}
\toprule
\multicolumn{1}{c}{\multirow{2}{*}{}} & \multicolumn{2}{c}{\textbf{Angle (Degrees)}}                  & \multicolumn{2}{c}{\textbf{Distance (Meters)}}                   \\ \cmidrule(lr){2-5} 
\multicolumn{1}{c}{}        & \multicolumn{1}{c}{\textbf{NN}} & \multicolumn{1}{c}{\textbf{Linear Regression}} & \multicolumn{1}{c}{\textbf{NN}} & \multicolumn{1}{c}{\textbf{Exponential Curve Fitting}} \\ \midrule
Mean Squared Error (MSE)          & \textbf{0.0467}        & 0.0468               & \textbf{0.0251}        & 0.6278                  \\ \midrule
$r^2$ score      & \textbf{0.9989}        & \textbf{0.9989}               & \textbf{0.9875}        & 0.9687                  \\ \bottomrule
\end{tabular}
\end{table}

\subsection{Camera Detection Results}

\begin{figure}[H]
\centering
\begin{subfigure}[b]{0.6\textwidth}
\includegraphics[width=\textwidth]{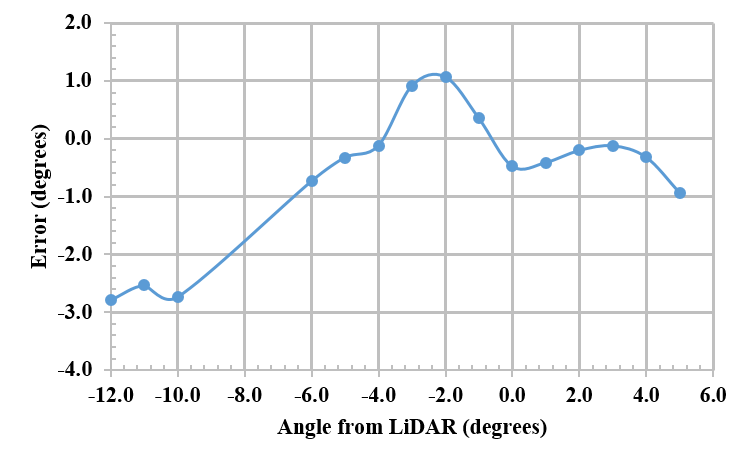}
\caption{}
\label{subfig:errora}
\end{subfigure}
\begin{subfigure}[b]{0.6\textwidth}
\includegraphics[width=\textwidth]{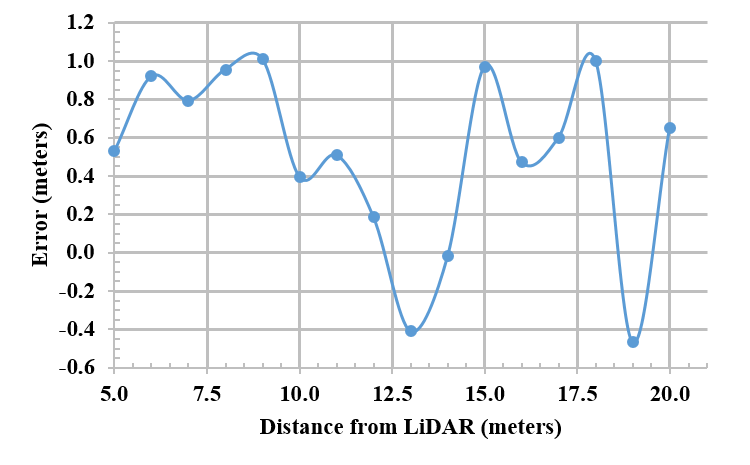}
\caption{}
\label{subfig:errord}
\end{subfigure}
\caption{(\textbf{a}) Angle error of the camera. (\textbf{b}) Distance error of the camera. Camera errors.}
\label{fig:error}
\end{figure}

In order to see how well the camera works on its own for reporting distance and angle for its detections, 400 testing images are processed. We compare the mapping results with the LiDAR detection, which is treated as the ground-truth. The results are shown in Figure~\ref{fig:error}. From Figure~\ref{fig:error}a, we~can see that the camera can fairly accurately predict the angle with a~maximum error around $3^{\circ}$, and this only happens near the image edges. On the other hand, the camera's distance prediction has relatively large errors, as shown in Figure~\ref{fig:error}b. Its maximum error is around 1 m, which happens when the distance is around 6 m, 9 m, 15 m and 18 m. 

\subsection{Fusion Results}
We also compare the true positive detections, false positive detections and false negative detections using only LiDAR data with linear SVM method, only camera image results with YOLO and the proposed fusion method. The results are shown in Tables~\ref{tbl:Fusion1} and \ref{tbl:Fusion2}.

In Tables~\ref{tbl:Fusion1} and \ref{tbl:Fusion2}, TPR, FPR and FNR are calculated by Equations (\ref{eq:TPR}) to (\ref{eq:FNR}), respectively, in~which TP is the number of true positives (the number of true beacons detected), TN is the number of true negatives (the number of non-beacons detected correctly), FP is the number of false positive (there is an object detected as a~beacon that is not a~beacon) and FN
 is the number of false negatives (there is a~beacon, but we have no detection).
 \begin{subequations}
 \begin{align}
 True \ Positive\ Rate\ (TPR) = \dfrac{TP}{TP + FN} \label{eq:TPR} \\
 False\ Positive \ Rate \ (FPR) = \dfrac{FP}{FP + TN} \label{eq:FPR} \\
 False\ Negative\ Rate\ (FNR) = \dfrac{FN}{TP + FN} \label{eq:FNR} 
 \end{align}
 \end{subequations} 

\begin{table}[H]
\centering
\caption{Fusion performance: 3 to 20-m range. The best results in {\textbf{bold.}}}
\label{tbl:Fusion1}
\begin{tabular}{cccc}
\toprule
       & \textbf{LiDAR Only} & \textbf{Camera Only} & \textbf{Fusion} \\ \midrule
True Positive Rate (TPR) & 93.90\% & \textbf{97.60}\% & \textbf{97.60}\% \\ \midrule
False Positive Rate (FPR) & 27.00\% & \textbf{0.00}\% & 6.69\% \\ \midrule
False Negative Rate (FNR) & 6.10\% & \textbf{2.40}\% & \textbf{2.40}\% \\ \bottomrule
\end{tabular}
\end{table} \unskip 
\begin{table}[H]
\centering
\caption{Fusion performance: 20 to 40-m range. The best results in {\textbf{bold.}}}
\label{tbl:Fusion2}
\begin{tabular}{ccc}
\toprule
       & \textbf{Camera Only} & \textbf{Fusion} \\ \midrule
True Positive Rate (TPR) & \textbf{94.80}\% & \textbf{94.80}\% \\ \midrule
False Positive Rate (FPR) & \textbf{0.00}\% & \textbf{0.00}\% \\ \midrule
False Negative Rate (FNR) & \textbf{5.20}\% & \textbf{5.20}\% \\ \bottomrule
\end{tabular}
\end{table} \unskip

\begin{figure}[H]
\centering
\caption{Benefit of using the camera and LiDAR fusion system.}
\label{fig:benefit}
\includegraphics[width=6in]{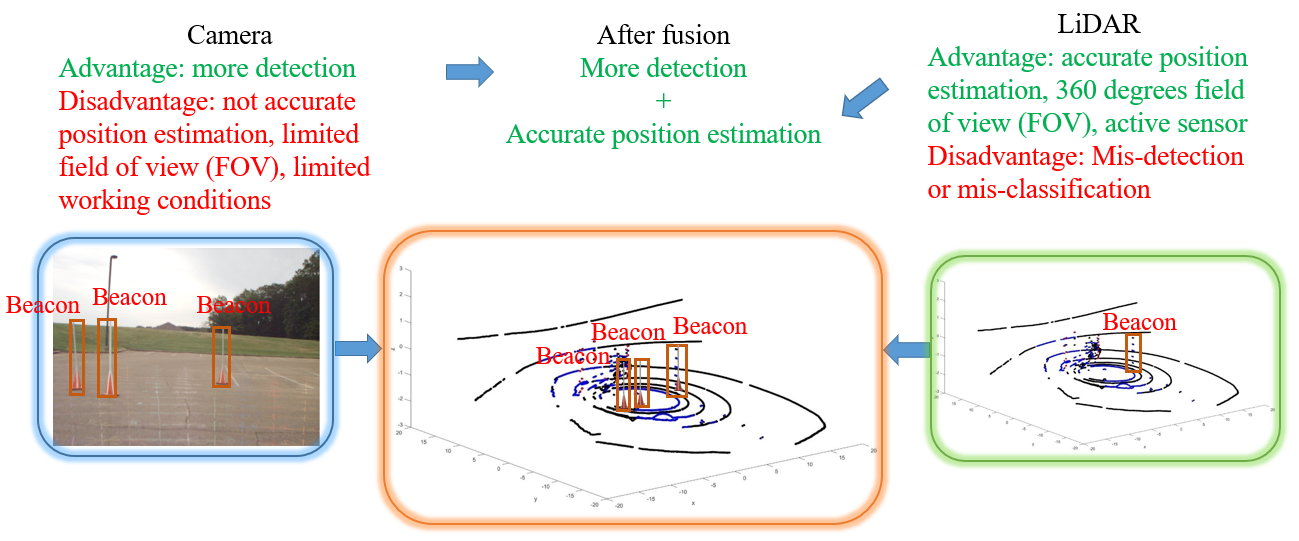}
\end{figure}

From these results, one can see that from the camera, there are more detections within its FOV up to 40 m, while its estimation of position, especially for distance, is not accurate. The camera, however, {can} fail more quickly in rain or bad weather, so we cannot always rely on the camera in bad weather~\citep{larochelle1998performance, park2012fog}. On the other hand, the LiDAR has accurate position estimation for each point, but the current LiDAR processing cannot reliably detect beyond about 20 m; and its false detection rate (including false positives and false negatives) is relatively high. After fusion, the results show that the system can detect beacons up to 40 m like the camera can, and the LiDAR can help more accurately estimate the position of these detections, which is important to the control system (which will need different control responses when there is a~beacon 20 m away versus the case where the beacon is 5 m away). Figure \ref{fig:benefit} shows the benefit of using the camera and LiDAR fusion system. 

\section{Conclusions and Future Work} \label{sec:conclusion}
In this paper, we proposed a~multi-sensor detection system that fuses camera and LiDAR detections for a~more accurate and robust beacon detection system. 
The proposed system has been designed as a~real-time industrial system for collision avoidance and tested with prototypes in various scenarios. 
The results show that fusion helps to obtain more accurate position and label information for each detection. It also helps to create detection beyond the range of LiDAR while within the range of the camera. 

In the future, we will try to implement multiple cameras or use wider FOV cameras to cover a wider FOV. We will also adjust the NN for more accurate prediction. We want to investigate using DL to learn to recognize beacons based on training data, without having to implement hand-crafted features. This approach can provide better results, but will be much more computationally demanding on the Jetson. {We experimented in low-light conditions and found that the camera detection system worked well until about 20 min after sunset, where the camera images were very dark and had little contrast. The system was able to detect beacons, but the localization (bounding boxes) were~poor. \mbox{In these cases}, we would rely on the LiDAR for detection, which really was not affected by low~light. Furthermore, we~want to construct an environmentally-controlled testing chamber where we can conduct experiments on camera degradation under known rain rates, amounts of fog and dust and~under low-light conditions.}

\vspace{6pt} 

\authorcontributions{{Data curation, Lucas Cagle, Pan Wei; Methodology, Pan Wei; Resources, Tasmia Reza and James Gafford;} Software, Lucas Cagle, Pan Wei; Writing – original draft, Pan Wei and John Ball } 



\conflictsofinterest{The authors declare no conflict of interest.} 

\reftitle{References}

\end{document}